\documentclass[lettersize,journal]{IEEEtran}
\usepackage{amsmath,amssymb,amsfonts}

\usepackage{algorithm}
\usepackage{algpseudocode}
\usepackage{algorithmicx}
\usepackage{array}
\usepackage{booktabs}
\usepackage{caption}
\usepackage{multirow}
\usepackage{subcaption}
\usepackage{textcomp}
\usepackage{stfloats}
\usepackage{url}
\usepackage{verbatim}
\usepackage{graphicx}
\usepackage{booktabs} % 导入三线表需要的宏包
\usepackage[pagebackref,breaklinks,colorlinks]{hyperref}
\usepackage[capitalize]{cleveref}
\usepackage{diagbox}
\usepackage{xcolor}
\usepackage{footnote}
\usepackage{bbding}
\begin{document}

\title{Efficient Learnable Collaborative Attention for Single Image Super-Resolution}

\author{Yi-Gang~Zhao
		Chao-Wei~Zheng, Jian-Nan~Su\IEEEauthorrefmark{1},
        Guang-Yong~Chen, \\
        and~Min~Gan,~\IEEEmembership{Senior Member,~IEEE,}% <-this % stops a space
\thanks{* Corresponding author (E-mail: sjn.fzu@gmail.com)}% <-this % stops a space
%\thanks{$\dagger$ https://github.com/}% <-this % stops a space
\thanks{Yi-Gang~Zhao, Jian-Nan~Su, Guang-Yong~Chen and Chao-Wei~Zheng are with the College of Computer and Data Science, Fuzhou University, Fuzhou 350108, China.}% <-this % stops a space
\thanks{Min~Gan is with the College of Computer Science and Technology, Qingdao University, Qingdao 266071, china, and also with the College of Computer and Data Science, Fuzhou University, Fuzhou 350108, China.}

%\thanks{C. L. Philip Chen is with the School of Computer Science and Engineering, South China University of Technology, Guangzhou 510641, China, and also with the College of Computer Science and Technology, Qingdao University, Qingdao 266071, China. E-mail: Philip.Chen@ieee.org}
}



\maketitle

\begin{abstract}
Non-Local Attention (NLA) is a powerful technique for capturing long-range feature correlations in deep single image super-resolution (SR). However, NLA suffers from high computational complexity and memory consumption, as it requires aggregating all non-local feature information for each query response and recalculating the similarity weight distribution for different abstraction levels of features. To address these challenges, we propose a novel Learnable Collaborative Attention (LCoA) that introduces inductive bias into non-local modeling. Our LCoA consists of two components: Learnable Sparse Pattern (LSP) and Collaborative Attention (CoA). LSP uses the \emph{k}-means clustering algorithm to dynamically adjust the sparse attention pattern of deep features, which reduces the number of non-local modeling rounds compared with existing sparse solutions. CoA leverages the sparse attention pattern and weights learned by LSP, and co-optimizes the similarity matrix across different abstraction levels, which avoids redundant similarity matrix calculations. The experimental results show that our LCoA can reduce the non-local modeling time by about 83\% in the inference stage. In addition, we integrate our LCoA into a deep Learnable Collaborative Attention Network (LCoAN), which achieves competitive performance in terms of inference time, memory consumption, and reconstruction quality compared with other state-of-the-art SR methods.

%Recently, researchers have utilized the Non-Local Attention (NLA) to explore the self-similarity of input images, significantly improving the performance of convolution-based Super-Resolution (SR) networks in image reconstruction. However, the NLA requires prohibitive computational costs and vast GPU memory occupation because it needs to aggregate all non-local features information for each query response. The further reason is that the NLA lacks certain ideal inductive biases to reduce computational complexity during non-local modeling. In this paper, we propose a novel Learnable Collaborative Attention (LCoA), which encodes inductive biases into the NLA in the form of sparsity and weight sharing to address this problem.  Specifically, we designed two essential modules: the Learnable Sparse Pattern (LSP) and the Collaborative Attention (RA). The LSP utilizes \emph{k}-means clustering algorithm to learn dynamic sparse attention patterns, while the CoA shares the sparse attention patterns and weights provided by the LSP to improve computational efficiency. Experimental results show that our LCoA has significant advantages over the NLA in terms of inference time and GPU memory consumption, while achieving better image reconstruction quality. Based on the LCoA, we construct a deep Learnable Collaborative Attention Network  (LCoAN), which achieves competitive performance in both quantitative metrics and perceptual quality compared to other state-of-the-art SISR models.
\end{abstract}


\section{Introduction}\label{sec_intro}
%Single image super-resolution (SISR), which aims to transform a given low-resolution (LR) image into a high-resolution (HR) image with better visual quality and finer details, enjoys a wide range of applications in the real world, such as video surveillance, satellite imaging and medical detection \cite{mahapatra2019image,rasti2016convolutional,shermeyer2019effects,wang2021wide,zhang2018longitudinally}. In general, the SISR problem is very challenging and inherently ill-posed, which leads to poor reconstruction performance of traditional methods \cite{glasner2009super,yang2010image}. With the rapid development of deep learning technology, the SISR models based on deep learning have been actively explored, among which the powerful feature representation and end-to-end training paradigm of convolution neural networks (CNNs) make it a promising approach to SISR. Thus, a flurry of CNNs-based models \cite{dong2014learning,kim2016accurate,lim2017enhanced,zhang2018residual} have been proposed and brought significant performance improvements on various SISR benchmarks.
The goal of single image super-resolution (SR) is to generate a high-resolution (HR) image with enhanced visual quality and more details from a given low-resolution (LR) image. SR has a wide range of real-world applications in fields such as video surveillance, satellite imaging and medical detection \cite{shermeyer2019effects,wang2021wide,zhang2018longitudinally}. However, SR is a very challenging and ill-posed problem, since one LR image can correspond to multiple HR images. Traditional methods for SR \cite{glasner2009super,yang2010image} often suffer from poor reconstruction performance due to their limited generalization ability.  In contrast, deep learning-based methods have demonstrated remarkable superiority over traditional methods for SR, owing to the powerful feature representation and end-to-end training paradigm of convolutional neural networks (CNNs). As a result, many very deep CNNs-based models \cite{kim2016accurate,lim2017enhanced,zhang2018residual} have been developed and achieved significant performance improvements on various image SR benchmarks.

%\cite{mahapatra2019image,rasti2016convolutional,shermeyer2019effects,wang2021wide,zhang2018longitudinally}

To enhance the image reconstruction ability of SR networks, it is not only essential to design deeper networks that can learn discriminative high-level features, but also to fully leverage the long-range feature correlations in intermediate layers that reflect the self-similarity of input images. Therefore, many researchers have started to explore the self-similarity of input images by using Non-Local Attention (NLA) \cite{wang2018non} and achieved satisfactory SR results \cite{dai2019second,liu2018non,xia2022efficient}. However, the NLA needs to aggregate information from all non-local features for the response of each query, which leads to prohibitive computational costs and vast GPU memory occupation. For the standard NLA, it lacks some desirable inductive biases to reduce the computational complexity during non-local modeling. Therefore,  this paper is dedicated to improving the computational efficiency of capturing long-range feature correlations within intermediate layers by incorporating reasonable inductive biases.

When exploring long-range feature correlations, existing SR models usually incorporate the NLA gradually into the network. However, this approach ignores the relationships between attention weights across different layers. For SR tasks, we observe an interesting phenomenon: the texture structure information of the image is stable in the network. This implies that the non-local relations at different abstraction levels have high correlation, and thus we can use this property as an inductive bias to collaboratively optimize the similarity matrix across different abstraction levels, greatly reducing the computational cost of using multiple NLAs. Additionally, in the experiment we found that  clustering the shallow features produces more accurate results than clustering the deep features. This may be due to the increasing degree of coupling between deep features as the network becomes deeper, which leads to the clustering results being unable to accurately reflect the similarity between textures in low-resolution images. To address this problem, we employ both shallow and deep features to collaboratively optimize the clustering results and attention weights.

Besides reducing the computational cost of using multiple NLAs through the aforementioned inductive bias, it is equally important to improve the computational efficiency of a single NLA itself. The main research direction is to use sparsity as an inductive bias during non-local modeling to improve efficiency. Current research attempts to constrain non-local operations within fixed sub-regions or use random projection local sensitive hashing to limit feature matching range \cite{guo2019star,zaheer2020big,kitaevreformer}. However, these fixed or data-independent sparse patterns either lose global modeling capability or face the challenge of high estimation variance. An ideal sparse pattern should be data-driven, capable of learning relevant sparse prior knowledge from rich training data to adapt to different images.

In this paper, we propose the Learnable Collaborative Attention (LCoA), which encodes inductive biases into non-local modeling. Thus, our LCoA not only preserves the ability to efficiently capture long-range feature correlations but also greatly reduces the computational cost and GPU memory occupation. Specifically, our LCoA consists of the Learnable Sparse Pattern (LSP) and the Collaborative Attention (CoA). The LSP relies on \emph{k}-means clustering to learn dynamic attention sparse patterns. Our strategy first assigns all non-local features to clusters, then only relevant features from the same cluster are considered for attention. To further improve computational efficiency, the sparsity pattern and attention weights learned by the LSP are co-optimized by all CoA. Experimental results on several popular datasets show that our LCoA has significant advantages over NLA in terms of inference time and GPU memory consumption, reducing by 82\% and 65\%, respectively. We also compared other efficient attention methods \cite{mei2021image,xia2022efficient} with our LCoA, and it showed outstanding advantages in both image reconstruction performance and computational efficiency. In summary, our contributions can be summarized in the following.
\begin{itemize}
\item{We proposed a novel Learnable Sparse Pattern (LSP) to capture self-similarity information for SR tasks. Compared with existing fixed or data-independent sparse patterns, our LSP exhibits competitive performance in exploring the self-similarity prior of images.}
\item{We designed a Collaborative Attention (CoA) mechanism that co-optimizes attention weights and clustering results to reduce computational costs while alleviating the issue where clustering of deep features fails to accurately reflect the similarity between textures in LR images.}
\item{A Learnable Collaborative Attention (LCoA) was proposed, which leverages LSP and CoA to induce learnable sparsity patterns and weight sharing biases into the process of non-local modeling.  The experimental results indicate that LCoA exhibits competitive performance in terms of computational efficiency and reconstruction results.}
\end{itemize}

\section{Related work}
Image super-resolution (SR) is a low-level computer vision task that aims to recover a high-resolution (HR) image from a low-resolution (LR) observation. It has various applications in security, surveillance, satellite, and medical imaging \cite{shermeyer2019effects,wang2021wide,zhang2018longitudinally}, and can also enhance the performance of other image processing or recognition tasks. In recent years, deep learning-based methods have achieved remarkable advances in SR, surpassing the traditional methods that rely on hand-crafted features or interpolation techniques \cite{glasner2009super,yang2010image}. Deep learning-based methods use a large amount of paired LR-HR images to train a deep neural network that learns a nonlinear mapping from LR to HR. These methods can be further divided into reconstruction-based methods and generative adversarial network (GAN)-based methods. Reconstruction-based methods optimize a pixel-wise loss function, such as mean squared error (MSE) or L1 norm, to minimize the difference between the output and the ground truth HR image. Some representative models include FSRCNN \cite{dong2016accelerating}, VDSR \cite{kim2016accurate}, EDSR \cite{lim2017enhanced}, RDN \cite{zhang2018residual}, and RCAN \cite{zhang2018image}. GAN-based methods introduce an adversarial loss to encourage the output to be more realistic and perceptually pleasing. The adversarial loss is computed by a discriminator network that tries to distinguish between real and fake HR images, while the generator network tries to fool the discriminator. Some examples of GAN-based models are SRGAN \cite{ledig2017photo}, ESRGAN \cite{wang2018esrgan}, CinCGAN \cite{yuan2018unsupervised}, and SROOE \cite{park2023perception}. 

Attention mechanisms \cite{fu2017look, hu2018squeeze, itti1998model, mnih2014recurrent} in SR networks, including channel attention and spatial attention, are widely used to enhance feature representation and extraction capabilities. Channel attention aims to adaptively recalibrate the feature responses across different channels according to their importance. Spatial attention aims to emphasize the salient regions or pixels in the feature maps according to their relevance, such as the representative non-local attention (NLA) \cite{wang2018non}. Some models combine both types of attention to achieve better performance, such as RCAN \cite{zhang2018image}. In deep learning-based image SR, NLA is widely used to explore self-similarity. For example, SAN \cite{dai2019second} used region-level non-local operations to capture long-range correlations in the entire feature map, which is suitable for low-level visual tasks. By introducing the cross-scale prior with in a powerful recurrent fusion cell, CSNLN \cite{mei2020image} can find more cross-scale feature correlations within a single LR image. ERN \cite{lan2020cascading} employs a dual global pathway structure that incorporates non-local operations to catch long-range dependencies from the the LR input. While the effectiveness of NLA in conjunction with SR networks has been proven, its application is limited by the high computational cost and the quadratic increase in size as the input image grows.

One of the acceleration methods is to exploit the sparsity in the attention matrices, which means that only a subset of the elements are non-zero and need to be computed. For example, sparse transformer \cite{child2019generating} employs a factorized attention mechanism that incorporates distinct sparse patterns tailored for various data types. BigBird \cite{zaheer2020big} incorporates random attention to approximate full attention and utilizes sparse encoder and decoder models However, these methods use static or fixed sparse patterns that may not adapt well to different input sequences. NLSA \cite{mei2021image} used spherical locality-sensitive hashing (LSH) to partition the input space into hash buckets with related features and compute only the attention within each bucket to reduce computational cost. While LSH may mistakenly scatter some related elements into different hash buckets, resulting in large estimation variances. ENLA \cite{xia2022efficient} decomposed the attention matrix by Gaussian random feature approximation and changed multiplication order to obtain linear complexity with respect to image size. But this unbiased approximation cannot guarantee that attention scores are non-negative, which may result in unstable and anomalous behavior. Inspired by the successful application of Routing Transformer \cite{roy2021efficient} in natural language processing, we utilized the \emph{k}-means clustering to construct an Learnable Sparse Pattern (LSP) block to learn relevant sparse prior knowledge from rich training data, adapting to different images and reducing computational complexity. In addition, we also leveraged the prior knowledge of texture structures in images to share the sparse model and attention weights of LSP in the network, further enhancing computational efficiency.

\begin{figure*}[!htbp]
	\hspace{-8mm}
	\centering
	\subfloat[]{
		\label{fig_network_arc}
		\includegraphics[width=1\linewidth]{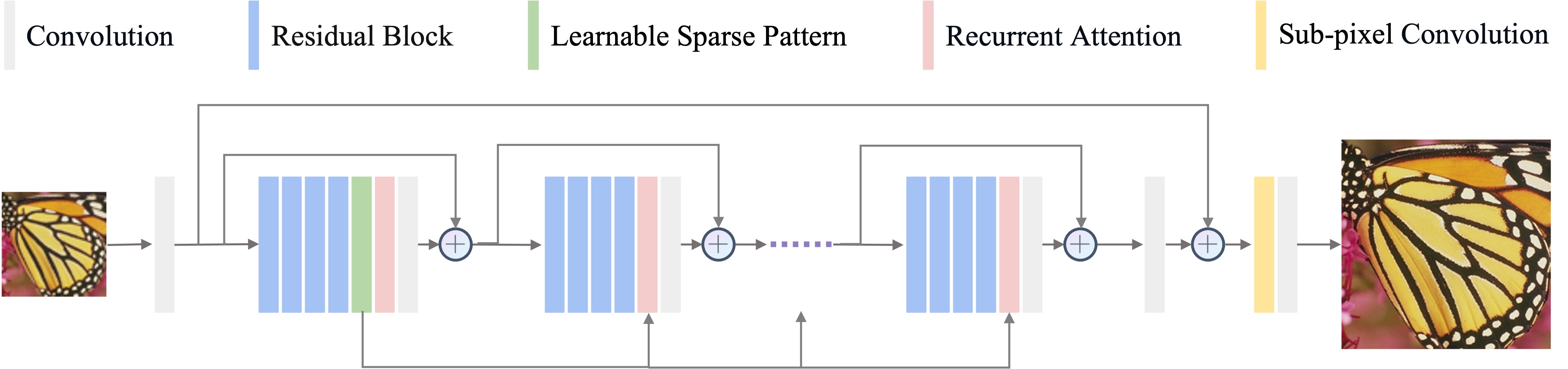}		
		\vspace{-1mm}
	}

	\subfloat[]{
		\label{fig_network_lsp}
		\includegraphics[width=0.55\linewidth]{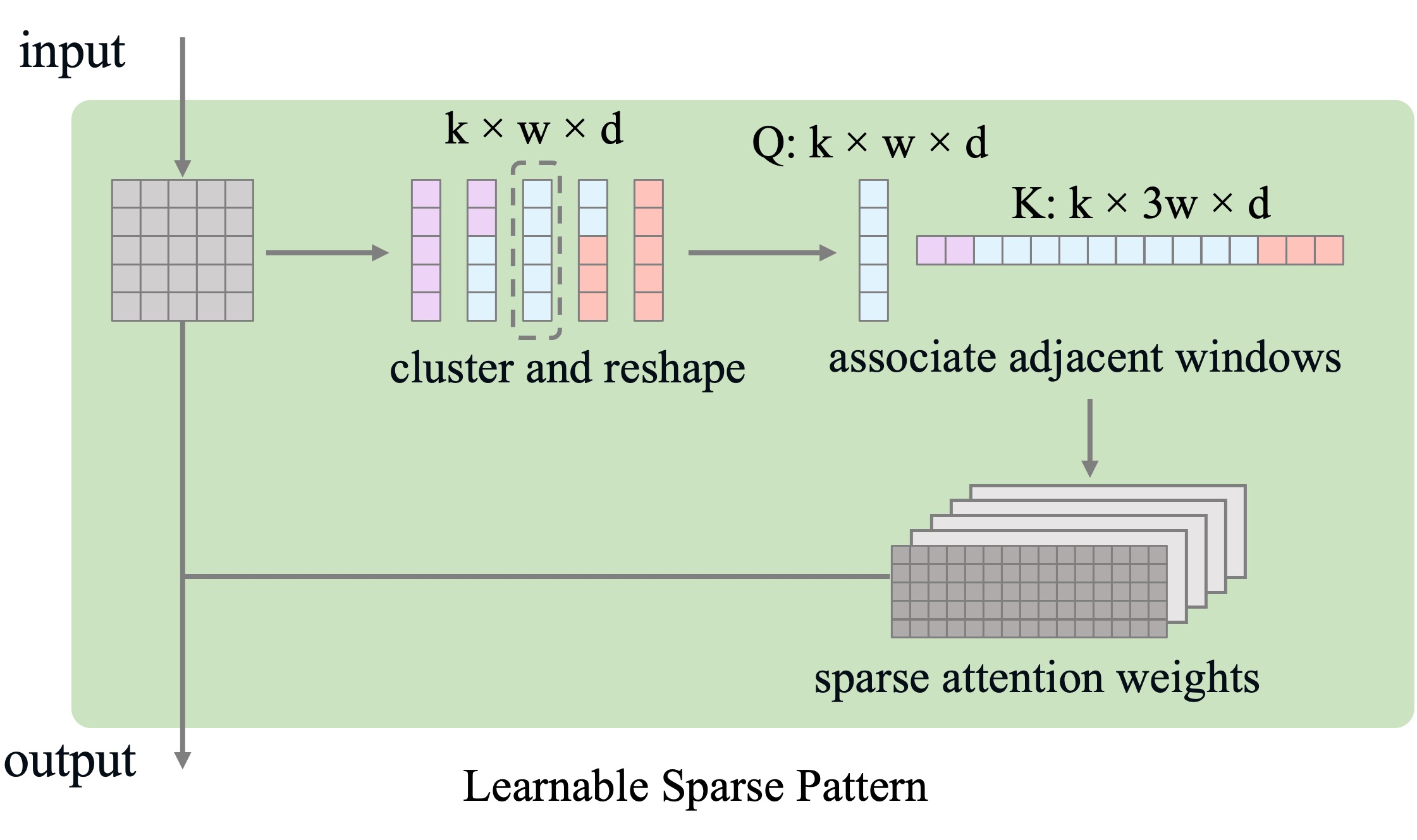}		
		\vspace{-3mm}
	}
	\subfloat[]{
		\label{fig_network_ca}
		\includegraphics[width=0.415\linewidth]{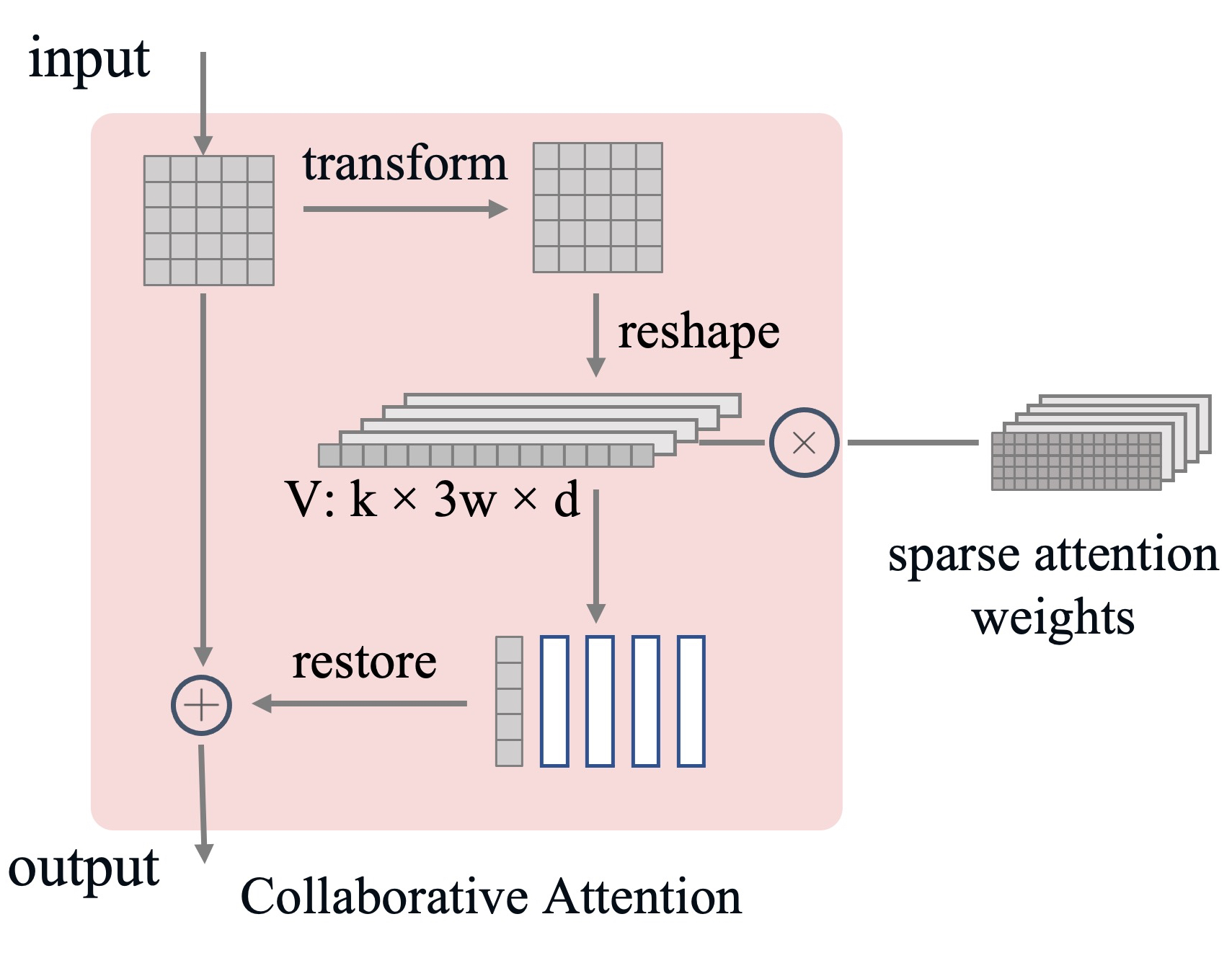}		
		\vspace{-3mm}
	}
	\caption{The structure of our Learnable Collaborative Attention Network (LCoAN). The LCoAN is built upon a deep residual network that incorporates Learnable Sparse Pattern (LSP) and the Collaborative Attention (CoA), and the sparsity pattern and attention weights optimized by the LSP are co-optimized by all CoA.}
	\label{fig_network}
\end{figure*}

\section{Methodology}
In this section, we first review the limitations of Non-Local Attention (NLA) \cite{wang2018non} in \cref{sec_review_nla}. Then the proposed Learnable Collaborative Attention, which mainly consists of the Learnable Sparse Pattern (LSP) and the Collaborative Attention (CoA), is detailed in \cref{sec_detail_lca}. LSP aims to learn dynamic attention patterns, and CoA further improves the computational efficiency of non-local modeling by weight sharing. Finally, our network architecture is introduced in \cref{sec_network_arch}.

\subsection{Reviewing Non-Local Attention} \label{sec_review_nla}
In general, the NLA explores global information by summarizing all non-local information from an input feature map. Specifically, given an input feature $X\in R^{h\times w\times c}$ with height $h$, width $w$ and $c$ channels, the NLA first applies three linear projections,
\begin{equation}
Q = W_q(X), \ K = W_k(X), \ V = W_v(X),
\label{eq_transform_x}
\end{equation}
where $Q \in R^{h \times w \times \hat{c}}$, $K \in R^{h \times w \times \hat{c}}$ and $V \in R^{h \times w \times \hat{c}}$ are referred to as queries, keys, and values, while $W_q$, $W_k$, and $W_v$ are three $1 \times 1$ convolutions. $\hat{c}$ is the channel number of the new embeddings. 

Next, $Q$, $K$, and $V$ are flattened to size $n \times \hat{c}$, where $n = h \times w$. Then, the similarity matrix $A \in R^{n \times n}$ is obtained by a matrix multiplication as
\begin{equation}
A = QK^{T}.
\label{eq_matrix_multiplication}
\end{equation}

Afterward, a normalization is applied to $A$ to get a unified similarity matrix as
\begin{equation}
A^{'} = \textsl{softmax}(A),
\label{eq_normalization}
\end{equation}
where the softmax operator over matrices denotes that the softmax function has been applied to each row. The unified similarity matrix $A^{'}$ may be interpreted as a matrix of weights in [0, 1] where $A^{'}_{ij}$ denotes how much query position $i$ at the output layer must pay attention to key position $j$ at the input layer. Given the unified similarity matrix $A^{'}$, the output of attention layer $O \in R^{n \times \hat{c}}$ is then computed simply as $A^{'}V$. In summary,  
\begin{equation}
O_i = \sum_{j}^{n}A^{'}_{ij}V_j.
\label{eq_normalization}
\end{equation}
The final output is given by
\begin{equation}
Y = W_o(O^T) + X,
\label{eq_final_output}
\end{equation}
where $W_o$ not only recovers the channel dimension from $\hat{c}$ to $c$ but also acts as a weighting parameter to adjust the importance of the non-local operation with respect to the original input $X$. In addition, the residual connection allows us to insert a new non-local block into any pre-trained model, without breaking its initial behavior.

% 最后一句可以考虑改一改
The NLA is potent to capture long-range feature correlations that are crucial for SR tasks. By inspecting the general computing flow, the NLA suffers from quadratic computation and memory requirements with respect to the image size, where instantiating the similarity matrix $A$ in \cref{eq_matrix_multiplication} dominates the cost. Our work is interested in addressing the inherent flaws of the NLA.

% 如等式2所示，相似矩阵事n*n维的。这导致基于NLA的模型，复杂度是二次的。我们的工作专注于降低计算复杂度。
% Despite its effectiveness, self-attention suffers from quadratic computation and memory requirements with respect to sequence length. 

\subsection{Learnable Collaborative Attention} \label{sec_detail_lca}
In this section, we will detail the proposed Learnable Collaborative Attention (LCoA), which encodes inductive biases into the NLA in the form of learnable sparsity and weight sharing. The LCoA comprises two parts, namely the Learnable Sparse Pattern (LSP) that enforces sparsity constraints, and the Collaborative Attention (CoA) that applies weight sharing to the NLA. Compared to the NLA, our LCoA significantly reduces the computational complexity to asymptotic linear, relative to the image size. We will now provide a comprehensive overview of LSP and CoA.

\subsubsection{Learnable Sparse Pattern} \label{sec_detail_lsp}
As previously mentioned, the prohibitive computational cost and vast GPU memory occupation have hindered the use of NLA in SR tasks. To address this issue, a common approach is to apply sparsity constraints to the NLA to improve computational efficiency. Specifically, for the sparse attention model, each query only attends to a subset of keys. We introduce the set $S_i$ as the subset of keys associated with the query at position $i$, namely:
\begin{equation}
O_i = \sum_{j \in S_i} A^{'}_{ij}V_{j}.
\label{eq_sparse_attention}
\end{equation}
%这里考虑增加可学习稀疏的优点

The set of all such keys defines the sparsity pattern $S = \{S_i | 1 \leq i \leq n\}$ of the input image. Previous works \cite{child2019generating,kitaevreformer,parmar2018image,zaheer2020big} have proposed fixed or data-independent sparsity patterns to guide the set $S$. Unlike other methods, the aim of this paper is to explore a more general form of attention sparsity, which learns sparse patterns from data and can be expressed as $S = f(X)$. To learn sparse patterns, we partition all non-local features using \emph{k}-means clustering, and only consider the relevant features from the same cluster for the attention mechanism. Specifically, we employ \emph{k}-means algorithm to cluster the keys $K$ and queries $Q$ onto the same set of centroid vectors $u = (u_1, ..., u_k) \in R^{k \times d}$. These centroid vectors are shared as model parameters across different images and can be learned online along with other parameters. After determining the clustering membership for queries and keys, the nearest centroids of $Q_i$ and $K_j$ are represented as $u(Q_i)$ and $u(K_j)$ respectively, both belonging to $u$. Therefore, we can define the sparse attention strategy as: 
\begin{equation}
O_i = \sum_{j:u(K_j) = u(Q_i)} A_{ij}V_{j}.
\label{eq_sparse_attention_strategy}
\end{equation}
The current query only attends to the keys that belong to the same cluster. In other words, the current query feature is only associated with a limited number of non-local features through its clustering.

In LSP, we treat queries and keywords as unit vectors and project them onto a unit sphere. This processing step means that:
\begin{equation}
\begin{aligned}
	|| Q_i - K_j||^2 & = ||Q_i||^2+||K_j||^2 - 2Q_i^TK_j \\
					 & = 2 - 2(Q^T_iK_j).
\end{aligned}
\label{eq_project_query_key}
\end{equation}
In addition, if $Q_i$ and $K_j$ belong to the same cluster center (i.e., $u(Q_i)=u(K_j)$), it can be known that there exist some $\varepsilon>0$ such that $||Q_i-u||,||K_i-u||<\varepsilon$. We can deduce the following conclusion:
%Observing \cref{eq_project_query_key}, it can be found that if $Q_i$ and $K_j$ belong to the same cluster center (i.e., $u(Q_i) = u(K_j)$), then it can be known that there is some $\varepsilon > 0$, such that $||Q_i - u||,||Q_i-u|| < \varepsilon$. This conclusion can be deduced from the triangle inequality theorem that:
\begin{equation}
	|| Q_i - K_j|| \leq ||Q_i - u|| + ||K_j - u|| < 2\varepsilon.
\label{eq_triangle_inequality}
\end{equation}
Combining \cref{eq_project_query_key} and \cref{eq_triangle_inequality}, we can obtain $Q^T_iK_j > 1 - 2\varepsilon^2$. Therefore, the attention weights of the keys that belong to the same cluster as the query is also relatively high.

%\begin{figure}[!t]
%  \centering
%  \includegraphics[width=\linewidth]{images/balance_window.jpg}
%   \caption{Addressing the issue of imbalanced feature quantities in clustering.}
%   \label{fig_balance_window}
%\end{figure}

During training, we use mini-batch \emph{k}-means algorithm to train the cluster centroids. Each cluster centroid $u$ is updated by an exponentially moving average of all the keys and queries assigned to it:
\begin{equation}
	u \leftarrow \lambda u + \frac{1 - \lambda}{2}\sum_{i:u(Q_i)=u}Q_i + \frac{1-\lambda}{2}\sum_{j:u(K_j)=u}K_j,
	\label{eq_update_centroids}
\end{equation}
where the decay parameter $\lambda = 0.999$. Ideally, the number of keys or queries assigned to each cluster centroid $u$ would be equal. However, this may not hold in practice because the number of features within each category tends to be imbalanced. This makes it impossible to perform parallel computation during network training. As shown in \cref{fig_network_lsp}, to overcome this challenge, we first sort the features according to their centroids and then partition the features into fixed-sized windows as the final clustering result. This strategy guarantees that all clusters have the same size, which is extremely important in terms of computational efficiency on parallel hardware like graphic cards. The drawback is that this allocation strategy may cause features from the same category to be assigned to different windows. Therefore, we allow attention to span across adjacent windows to effectively mitigate this drawback.

%\begin{figure}[!t]
%  \centering
%  \includegraphics[width=0.7\linewidth]{images/recurrent_att.jpg}
%   \caption{The diagram of Recurrent Attention.}
%   \label{fig_recurrent_att}
%\end{figure}

\subsubsection{Collaborative Attention} \label{sec_detail_ca}
As discussed in \cref{sec_intro}, the texture structure information of the image is stable across the network, thus we can leverage this property to collaboratively optimize the similarity matrix across different abstraction levels. We share attention weights in the network and reuse hidden states from shallow to deep layers. Weight sharing reduces redundant computations and also decreases memory usage, because some hidden states are stored in the same block of memory.

Specifically, we first use LSP to calculate the sparse attention weight matrix $A_{s}$ on the shallow features $X_1$ of the network. This process can be formally defined as:
\begin{equation}
	A_s = LSP(X_1).
	\label{eq_caculate_sparse_attention}
\end{equation}

Then, the proposed CoA shares weights to capture the long-range feature correlations in intermediate layers. The output of the $m$-th CoA in the network $O_m$ is
\begin{equation}
	O_m = X_m + \beta A_sW_m(X_m),
	\label{eq_collaborative_attention}
\end{equation}
where a linear transformation is applied to the input feature $X_m$, and $\beta$ is a scaling parameter. 

The processing flow of CoA is illustrated in \cref{fig_network_ca}. First, the input features are linearly transformed only once, and then rearranged according to the indices of the sparse attention. Next, long-range feature correlations are modeled by shared sparse attention weights to improve the efficiency of non-local modeling. Finally, rearrange the output back into its original position according to the same index.

% As shown in \cref{fig_network_ca},
Since the weights are provided by the LSP, the two linear transformations (i.e., Q and K) in the NLA can be ignored, which also reduces the number of model parameters and memory usage during inference. In addition, sharing parameters can reduce the complexity of the model and make the network easier to train. For example, in our experiments, we saw that CoA not only greatly reduced inference time but also improved image reconstruction results on some datasets.

%\begin{figure*}[!t]
%  \centering
%  \includegraphics[width=\linewidth]{images/network2.jpg}
%   \caption{The structure of our Learnable Recurrent Attention Network.}
%   \label{fig_network}
%\end{figure*}

%\begin{figure*}
%	
%    \begin{minipage}[]{0.95\linewidth}
%        \centering
%        	\begin{subfigure}[]{\linewidth}
%				\centering
%				\includegraphics[width=\linewidth]{images/lcan.jpg}
%				\vspace{-8mm}
%        		\caption*{(a)}
%			\end{subfigure}
%			\label{fig_network_arc}
%    \end{minipage}
%    
%    \begin{minipage}[]{0.55\linewidth}
%        \centering
%        	\begin{subfigure}[]{\linewidth}
%        		\label{fig_network_lsp}
%				\centering
%				\includegraphics[width=\linewidth]{images/lsp.jpg}
%				\vspace{-8mm}
%        		\caption*{(b)}
%			\end{subfigure}
%    \end{minipage}
%    \hspace{5mm}
%    \begin{minipage}[]{0.4\linewidth}
%        \centering
%        	\begin{subfigure}[]{\linewidth}
%        	    \label{fig_network_ca}
%				\centering
%				\includegraphics[width=\linewidth]{images/ca.jpg}
%				\vspace{-8mm}
%        		\caption*{(c)}
%			\end{subfigure}
%    \end{minipage}
%    \centering
%    
%%\vspace{2mm}
%   \caption{The structure of our Learnable Collaborative Attention Network.}
%   \label{fig_network}
%\end{figure*}
%

\subsection{Network Architecture} \label{sec_network_arch}
%The overall architecture of our network is depicted in Fig. 3. A deep residual network with LSP and CoA builds LCoAN. Specifically, our LCoAN mainly consists of four parts: LR feature extraction, learnable sparse pattern, deep feature fusion and HR image reconstruction. Then, the learnable sparse pattern is trained on the shallow feature to explore the sparse prior of natural images and provide a sparse attention weight matrix. In the deep feature fusion stage, we construct the basic modules of the network with residual modules and recurrent attention, and the recurrent attention shares learnable sparse weights. At the end of the network, we apply a convolutional layer with a filter with 3 trainable parameters to reconstruct the output image.
The overall architecture of our network is depicted in \cref{fig_network_arc}. A deep residual network with LSP and CoA builds the deep Learnable Collaborative Attention Network (LCoAN). Specifically, our LCoAN mainly consists of three parts: LR feature extraction, deep feature aggregation, and HR image reconstruction. Firstly, a single convolutional layer is used to extract shallow features from the LR input. In the deep feature fusion stage, we construct the basic modules of the network with residual modules and CoA, and the CoA shares learnable sparse weights. The LSP is trained on the shallow layers of the network to explore the sparse prior of natural images and provide a sparse attention weight matrix. At the end of the network, we apply a convolutional layer with 3 trainable parameters to reconstruct the output image.

\section{Experiments}\label{sec:experiments}
\subsection{Setup}
\noindent\textbf{Datasets and Metrics.} We followed previous works \cite{dai2019second, lim2017enhanced, mei2021image} and used 800 images from DIV2K \cite{timofte2017ntire} as our training dataset. To test the effectiveness of our approach, we evaluated its performance on 5 standard benchmarks: Set5 \cite{bevilacqua2012low}, Set14 \cite{zeyde2010single}, BSD100 \cite{martin2001database}, Urban100 \cite{huang2015single}, and Manga109 \cite{matsui2017sketch}. All of the SR results were evaluated using the PSNR and SSIM metrics on the Y channel of the transformed YCbCr space.

\noindent\textbf{Implementations.} We integrate the proposed Collaborative Attention and Learnable Sparse Pattern into the residual backbone network, and name it as the deep Learnable Collaborative Attention Network (LCoAN). All intermediate features of our network have 128 channels. To convert deep features into a 3-channel RGB image, the last convolution layer in our LCoAN has 3 filters. We set all convolutional kernel sizes to $3 \times 3$.

\noindent\textbf{Training Details.} We randomly crop $48 \times 48$ patches from the training examples during training and create a mini-batch consisting of 16 images. To augment the dataset, we apply random rotations of 90, 180, and 270 degrees, as well as horizontal flipping. We optimize the model using the ADAM optimizer \cite{kingad2015methodforstochasticoptimization} with $\beta_1 = 0.9$, $\beta_2 = 0.999$, and $\epsilon = 10^{-8}$. For the scale factor $\times 2$, we set the initial learning rate to $10^{-4}$, which is halved after 300 epochs. The final model is obtained after 1500 epochs. Furthermore, we train the model parameters for scale factors $\times 3$ and $\times 4$ using the pre-trained $\times 2$ network, with the learning rate of $10^{-4}$ reduced to half every 100 epochs until the training stops at 500 epochs. Our model is implemented using PyTorch and trained on Nvidia 3090 GPUs.

\begin{table}[!htbp]
\centering
\caption{Ablation study of LCoA (comprising LSP and CoA). The best result is highlighted.}
\label{tab_ablation_core}
\vspace{1mm}
\begin{tabular}{|c|c|c|c|c|c|}
\hline
\textbf{Case} & \textbf{1}  & \textbf{2}     & \textbf{3}  & \textbf{4} & \textbf{5 (ours)}  \\ \hline \hline
Backbone           & \Checkmark     & \Checkmark     & \Checkmark  & \Checkmark & \Checkmark  \\ \hline
NLA           & \XSolidBrush     & \Checkmark     & \XSolidBrush  & \XSolidBrush & \XSolidBrush  \\ \hline
LSP    & \XSolidBrush  & \XSolidBrush     & \Checkmark     & \XSolidBrush & \Checkmark     \\ \hline
LCA with CoA  & \XSolidBrush     & \XSolidBrush  & \XSolidBrush     & \Checkmark    & \Checkmark  \\ \hline 
PSNR (dB) 	& 33.52 & 33.55 & 33.58 & 33.58 & \textbf{33.64} \\ \hline \hline
Memory (MB) 	& 2846 & 15668 & 14786 & 5724 & \textbf{5518} \\ \hline
Time (s) 	& 32 & 272 & 133 & 69 & \textbf{48} \\ \hline
\end{tabular}
%\footnotesize{}
\end{table}

\subsection{Ablation Study}
In this section, we conduct controlled experiments and analyze the results on benchmark datasets to investigate the effectiveness of our Learnable Collaborative Attention (LCoA) mechanism. Our baseline model is built on a residual backbone with 10 Feature Aggregation Units (FAUs), and we replace the attention variants in each FAU to evaluate their impact. These experiments all run for $5 \times 10^4$ iterations.

%\begin{figure}[!htbp]
%	\hspace{-8mm}
%	\centering
%	\subfloat[]{
%		\label{fig_plot_u100_time}
%		\includegraphics[width=0.5\linewidth]{images/plot_u100_time.jpg}}
%	\subfloat[]{
%		\label{fig_plot_u100_memory}
%		\includegraphics[width=0.5\linewidth]{images/plot_u100_memory.jpg}}
%	\caption{The PSNR results in (a) and (b) are tested on Urban100 with scale factor 2. (a) Inference time comparisons. (b) Memory consumption comparisons.}
%	\label{fig_plot_u100}
%\end{figure}

% 消融实验图
\begin{figure}[!t]
  \centering
  \includegraphics[width=\linewidth]{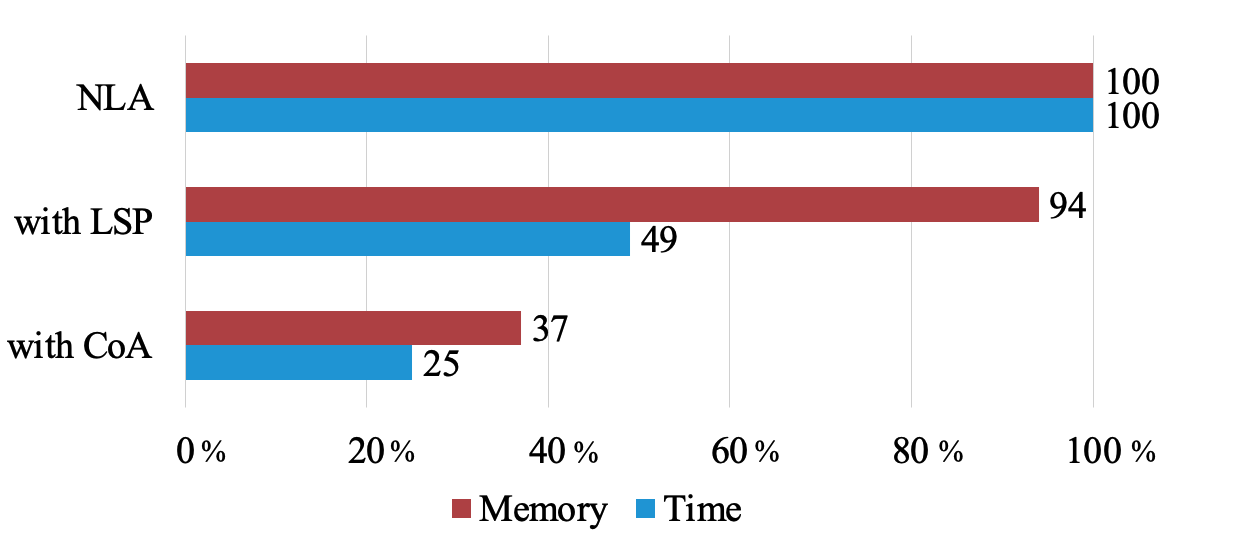}
  \caption{The impact of the proposed LSP and CoA on memory consumption and inference time on Urban100 ($\times 2$).}
  \label{fig_time_and_memory}
\end{figure}

%\begin{table}[!htbp]
%\centering
%\caption{Ablation study of LCoA (comprising LSP and CoA). The best result is highlighted.}
%\label{tab_ablation_core}
%\vspace{1mm}
%\begin{tabular}{|c|c|c|c|c|c|}
%\hline
%\textbf{Case} & \textbf{1}  & \textbf{2}     & \textbf{3}  & \textbf{4} & \textbf{5 (ours)}  \\ \hline \hline
%Backbone           & \Checkmark     & \Checkmark     & \Checkmark  & \Checkmark & \Checkmark  \\ \hline
%NLA           & \XSolidBrush     & \Checkmark     & \XSolidBrush  & \XSolidBrush & \XSolidBrush  \\ \hline
%LSP    & \XSolidBrush  & \XSolidBrush     & \Checkmark     & \XSolidBrush & \Checkmark     \\ \hline
%LCA with CoA  & \XSolidBrush     & \XSolidBrush  & \XSolidBrush     & \Checkmark    & \Checkmark  \\ \hline 
%PSNR (dB) 	& 33.52 & 33.55 & 33.58 & 33.58 & \textbf{33.64} \\ \hline \hline
%Memory (MB) 	& 2846 & 15668 & 14786 & 5724 & \textbf{5518} \\ \hline
%Time (s) 	& 32 & 272 & 133 & 69 & \textbf{48} \\ \hline
%\end{tabular}
%%\footnotesize{}
%\end{table}

\subsubsection{Effects of LSP and CoA}
To evaluate the effects of Learnable Sparse Pattern (LSP) and Collaborative Attention (CoA) in FAU, we tested on Set14 and Urban100 datasets respectively to compare the reconstruction quality and computational efficiency, and obtained the following experimental results. In case 1, we tested the residual backbone network without any attention modules in its FAUs; In case 2, we add the Non-Local Attention (NLA) to the FAU. In case3, we impose learnable sparsity constraints on non-local modeling through our LSP. In case4, the attention weights are calculated by the first non-local attention module in the network and shared by all subsequent attention. In case 5, we incorporated the proposed LCoA, comprising LSP and CoA, into the FAU.

% As shown in \cref{tab_ablation_core}, utilizing the NLA to capture the feature correlations of the intermediate layer in the network can improve the representational capability of the convolutional network and further enhance the performance of image reconstruction. Observing the experimental results of Cases 3 and 4, we found that both LSP and CoA contribute to the improvement of PSNR. This suggests that adding inductive biases of sparsity and weight sharing to NLA can effectively enhance the image reconstruction performance. Because NLA integrates a large number of irrelevant features during the modeling process, LSP filters out noise features by applying sparse constraints, thereby improving image reconstruction quality. In addition, the receptive field of deep-layer features tends to be global, which causes deep-layer features to be coupled with each other and difficult to accurately calculate their similarity. CoA calculates the attention weights of features in the shallow layers of the network and shares them with the deep layers, solving these problems and improving image reconstruction performance. Moreover, in Case 5 where LSP and CoA are combined, the results are much better than when used separately.
% Using non-local attention to explore the self-similarity of images can further enhance the image reconstruction performance of the network. 

\begin{figure}[!htbp]
  \centering
  \includegraphics[width=\linewidth]{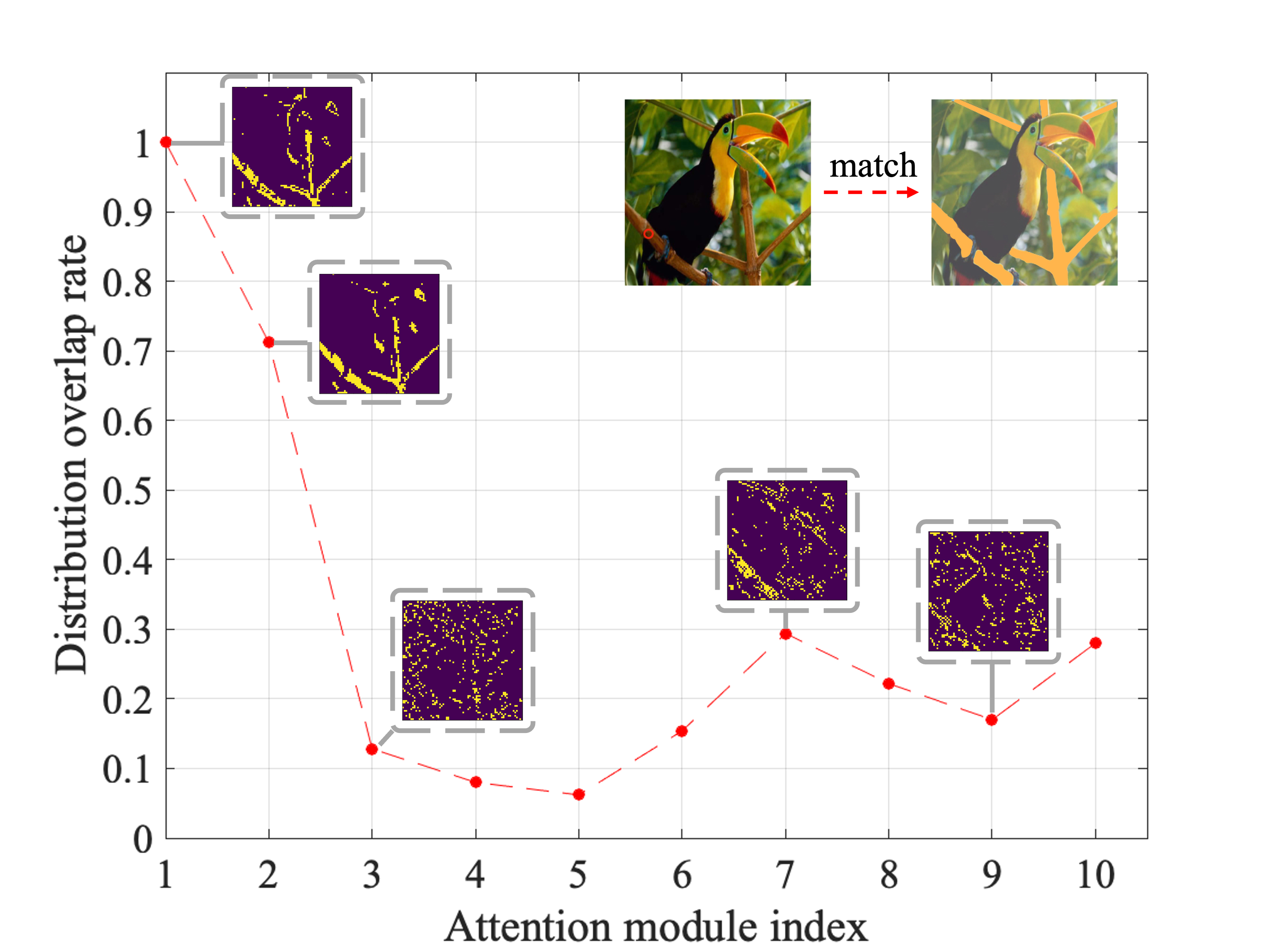}
  \caption{The distribution overlap rate between shallow attention maps and deep attention maps. We can observe that the distribution of shallow attention maps mainly focuses on areas similar to the query texture, while deep attention maps tend to be more randomly distributed.}
  \label{fig_distribute_rate}
\end{figure}

Comparing the results of Cases 1 and 2 in \cref{tab_ablation_core}, we can conclude that using NLA to explore the self-similarity of images can further enhance the image reconstruction performance of the network. Observing the experimental results of Cases 3 and 4, we found that both LSP and CoA contribute to the improvement of PSNR. This suggests that adding inductive biases of weight sharing and learnable sparsity to non-local modeling can effectively enhance the image reconstruction performance. Because LSP can filter out noise features by applying learnable sparse constraints to improve image reconstruction quality. In addition, deep-layer features are often coupled with each other, which may cause the deep attention maps of the network fail to accurately reflect the texture similarity in LR images, as shown in \cref{fig_distribute_rate}. By calculating the attention weights of shallow features in the network and sharing them with the deep layers, CoA mitigates the aforementioned issues, thereby improving the performance of image reconstruction. Moreover, in the case 5, combining LSP and CoA produces much better results than using them separately. From the \cref{tab_ablation_core}, it can be seen that the biggest advantage of LSP and CoA is that they can improve efficiency without compromising image reconstruction performance. The \cref{fig_time_and_memory} shows the advantages of LSP and CoA in reducing memory consumption and inference time based on the \cref{tab_ablation_core} and with NLA as the benchmark. On the challenging Urban100 dataset, both LSP and CoA can significantly reduce the inference time of attention. Specifically, LSP applied sparse constraints to non-local modeling reducing the inference time by approximately 50\%. And CoA implemented weight sharing of non-local modeling reducing the inference time by about 75\%. We obtained similar conclusions by comparing memory consumption.

\begin{figure}[!t]
	\hspace{-6mm}
	\centering
	\subfloat[]{
		\label{fig_plot_kmeans_rounds}
		\includegraphics[width=0.5\linewidth]{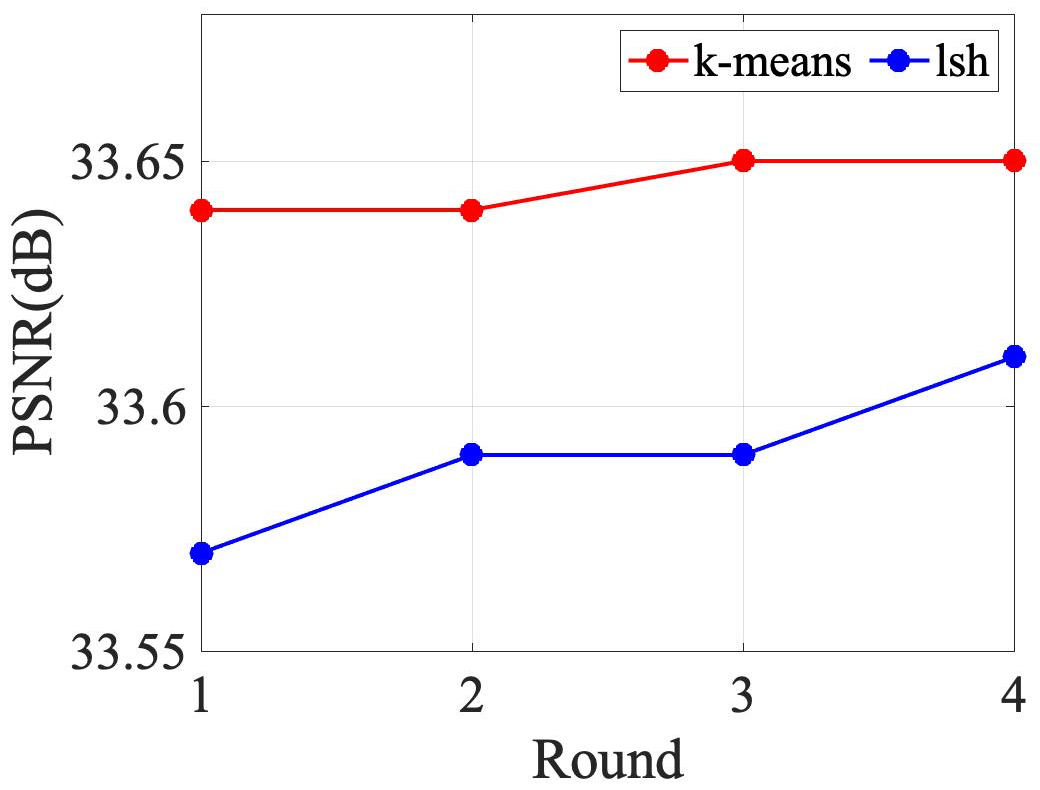}}
	\subfloat[]{
		\label{fig_plot_att}
		\includegraphics[width=0.5\linewidth]{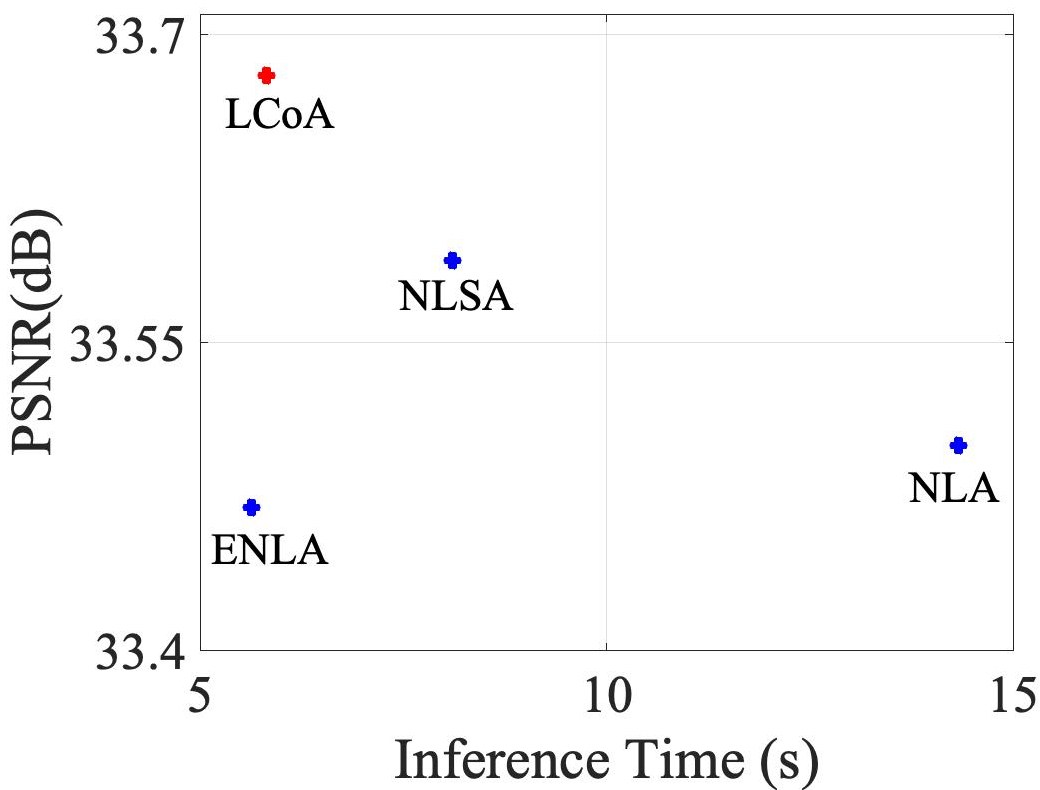}}
	\caption{Ablation experiments conducted on Set14 with scale factor 2 to explore the advantages of LCoA. (a) The PSNR results of replacing \emph{k}-means with LSH. (b) The comparison result of different attentions in terms of performance and efficiency.}
	\label{fig_plot_compare_lca}
\end{figure}

\subsubsection{Advantages of LCoA}
%To demonstrate the efficiency of LCoA, we built two SR networks that consist of only ten attention modules, using NLA and LCoA as the attention modules respectively, named as NLA-Net and LCoA-Net. As shown in \cref{tab_ablation_attention}, compared with NLA-Net, LCoA-Net saved about 83\% of the inference time, which fully reflected the superiority of LCoA. 
To demonstrate the efficiency of LCoA, we built two super-resolution networks that consist of only ten attention modules, using NLA and LCoA as the attention modules respectively, named as NLA-Net and LCoA-Net. We evaluated them on the Manga109 and found that LCoA-Net's inference time was only 20 seconds, while NLA-Net required 116 seconds, demonstrating an approximately 83\% reduction in non-local modeling time during the inference stage with our LCoA method. In addition, previous works have proposed sparse attention methods based on local sensitive hashing (LSH), such as NLSA \cite{mei2021image}. However, LSH, as a data-independent method, cannot learn sparse priors from training data, leading to poor generalization and large estimation variances. Our proposed LCoA utilizes \emph{k}-means clustering to group features and provides learnable sparse constraints, resulting in better robustness and performance. When we replaced \emph{k}-means with LSH in LCoA, experimental results showed that \emph{k}-means only needed one round of clustering to obtain accurate sparse results, leading to better performance in image reconstruction, as shown in \cref{fig_plot_kmeans_rounds}. Although LSH can improve the robustness of sparsity by increasing the number of hash rounds, its performance is still inferior to that of \emph{k}-means with a single round of clustering. These results emphasize the advantages of \emph{k}-means in providing more accurate and robust sparse results. In addition, to demonstrate the superiority of our LCoA method over other representative state-of-the-art attention methods, we conducted the following experiments. We replaced our LCoA with NLA \cite{wang2018non}, NLSA \cite{mei2021image}, and ENLA \cite{xia2022efficient} in the network and compared their performance in terms of PSNR and inference time. For a fair comparison, all attentions were trained with the same L1 loss function. As shown in \cref{fig_plot_att}, our LCoA outperformed other attention methods in PSNR and had a competitive performance in inference time.

%\begin{figure}[!t]
%  \centering
%  \includegraphics[width=\linewidth]{images/plot_att2.jpg}
%   \caption{Comparison of different attentions in terms of performance and efficiency on Set14 (×2).}
%   \label{fig_plot_att}
%\end{figure}

\begin{table*}[!htbp]
\begin{center}
\caption{Quantitative results on SR benchmark datasets. Best and second best results are \textbf{highlighted} and \underline{underlined}.}
\label{tab:x2_x3_x4_psnr_ssim}
%\resizebox{\linewidth}{6cm}
%{
\scalebox{1.1}{
\begin{tabular}{|c|c|c|cc|cc|cc|cc|cc|}
\hline
\multirow{2}{*}{Method} &
  \multirow{2}{*}{Scale} &
  \multirow{2}{*}{Param} &
%  \multirow{2}{*}{Time(s)} &
  \multicolumn{2}{c|}{Set5} &
  \multicolumn{2}{c|}{Set14} &
  \multicolumn{2}{c|}{B100} &
  \multicolumn{2}{c|}{Urban100} &
  \multicolumn{2}{c|}{Manga109} \\ \cline{4-13} 
   &
   &
   &
%   &
  \multicolumn{1}{c|}{PSNR} &
  SSIM &
  \multicolumn{1}{c|}{PSNR} &
  SSIM &
  \multicolumn{1}{c|}{PSNR} &
  SSIM &
  \multicolumn{1}{c|}{PSNR} &
  SSIM &
  \multicolumn{1}{c|}{PSNR} &
  SSIM \\ \hline

\begin{tabular}[c]{@{}c@{}}Bicubic\\ FSRCNN\\ VDSR\\ LapSRN\\ EDSR\\ RDN\\ RCAN\\ SAN\\ NLSN\\ DRLN \\ PACN\\ TAN\end{tabular} &
  \begin{tabular}[c]{@{}c@{}}$\times 2$\\ $\times 2$\\ $\times 2$\\ $\times 2$\\ $\times 2$\\ $\times 2$\\ $\times 2$\\ $\times 2$\\ $\times 2$\\ $\times 2$\\ $\times 2$\\ $\times 2$\end{tabular} &
    \begin{tabular}[c]{@{}c@{}}-\\ 57K\\ 12K\\ 812K\\ 40.73M\\ 22.12M\\ 15.44M\\ 15.71M\\ 41.80M\\ 34.43M\\ 15.32M\\ 16.12M\end{tabular} &
%      \begin{tabular}[c]{@{}c@{}}$\times 2$\\ $\times 2$\\ $\times 2$\\ $\times 2$\\ $\times 2$\\ $\times 2$\end{tabular} &
  \multicolumn{1}{c|}{\begin{tabular}[c]{@{}c@{}}33.66\\37.05\\37.53\\37.52\\ 38.11\\38.24\\ 38.27\\ 38.31\\ \textbf{38.34}\\ 38.27\\ 38.27\\ 38.27\end{tabular}} &
  \begin{tabular}[c]{@{}c@{}}0.9299\\0.9560\\0.9590\\0.9591\\ 0.9602\\ 0.9614\\ 0.9614\\ \textbf{0.9620}\\ 0.9618\\ 0.9616\\ 0.9613\\ 0.9614\end{tabular} &
  \multicolumn{1}{c|}{\begin{tabular}[c]{@{}c@{}}30.24\\32.66\\33.05\\33.08\\ 33.92\\ 34.01\\ 34.12\\ 34.07\\ 34.08\\ \textbf{34.28}\\ 34.03\\ 34.15\end{tabular}} &
  \begin{tabular}[c]{@{}c@{}}0.8688\\0.9090\\0.9130\\0.9130\\ 0.9195\\ 0.9212\\ 0.9216\\ 0.9213\\ \underline{0.9231}\\ \underline{0.9231}\\ 0.9211\\ 0.9219\end{tabular} &
  \multicolumn{1}{c|}{\begin{tabular}[c]{@{}c@{}}29.56\\31.53\\31.90\\31.08\\ 32.32\\ 32.34\\ 32.41\\ 32.42\\ \underline{32.43}\\ 32.39\\ 32.42\\ \textbf{32.44}\end{tabular}} &
  \begin{tabular}[c]{@{}c@{}}0.8431\\0.8920\\0.8960\\0.8950\\ 0.9013\\ 0.9017\\ 0.9027\\ 0.9028\\ 0.9027\\ 0.9028\\ 0.9025\\ 0.9027\end{tabular} &
  \multicolumn{1}{c|}{\begin{tabular}[c]{@{}c@{}}26.88\\29.88\\30.77\\30.41\\ 32.93\\ 32.89\\ 33.34\\ 33.10\\ \textbf{33.42}\\ \underline{33.37}\\ 33.18\\ 33.35\end{tabular}} &
  \begin{tabular}[c]{@{}c@{}}0.8403\\0.9020\\0.9140\\0.9101\\ 0.9351\\ 0.9353\\ 0.9384\\ 0.9370\\ \textbf{0.9394}\\ 0.9390\\ 0.9375\\ 0.9385\end{tabular} &
  \multicolumn{1}{c|}{\begin{tabular}[c]{@{}c@{}}30.80\\36.67\\37.22\\37.27\\ 39.10\\ 39.18\\ 39.44\\ 39.32\\ \underline{39.59}\\ 39.58\\ 39.44\\ 39.47\end{tabular}} &
  \begin{tabular}[c]{@{}c@{}}0.9339\\0.9710\\0.9750\\0.9740\\ 0.9773\\ 0.9780\\ 0.9786\\ 0.9792\\ \textbf{0.9789}\\ 0.9786\\ 0.9788\\ 0.9787\end{tabular} \\ \hline
\begin{tabular}[c]{@{}c@{}}Backbone\\ LCoAN\end{tabular} &
  \begin{tabular}[c]{@{}c@{}}$\times 2$\\ $\times 2$\end{tabular} &
    \begin{tabular}[c]{@{}c@{}}14.03M\\ 15.67M\end{tabular} &
%      \begin{tabular}[c]{@{}c@{}}$\times 2$\\ $\times 2$\end{tabular} &
  \multicolumn{1}{c|}{\begin{tabular}[c]{@{}c@{}} 38.23 \\ \textbf{38.34}\end{tabular}} &
  \begin{tabular}[c]{@{}c@{}} 0.9613 \\ \underline{0.9619}\end{tabular} &
  \multicolumn{1}{c|}{\begin{tabular}[c]{@{}c@{}} 34.01 \\ \underline{34.19}\end{tabular}} &
  \begin{tabular}[c]{@{}c@{}} 0.9203 \\ \textbf{0.9233}\end{tabular} &
  \multicolumn{1}{c|}{\begin{tabular}[c]{@{}c@{}} 32.34 \\ 32.42\end{tabular}} &
  \begin{tabular}[c]{@{}c@{}} 0.9019 \\ \textbf{0.9030}\end{tabular} &
  \multicolumn{1}{c|}{\begin{tabular}[c]{@{}c@{}} 32.87 \\ \underline{33.37}\end{tabular}} &
  \begin{tabular}[c]{@{}c@{}} 0.935 \\ \underline{0.9391}\end{tabular} &
  \multicolumn{1}{c|}{\begin{tabular}[c]{@{}c@{}} 39.17 \\ \textbf{39.61}\end{tabular}} &
  \begin{tabular}[c]{@{}c@{}} 0.9782 \\ \textbf{0.9789}\end{tabular} \\ \hline
\begin{tabular}[c]{@{}c@{}}Bicubic\\ FSRCNN\\ VDSR\\ LapSRN\\ EDSR\\ RDN\\ RCAN\\ SAN\\ NLSN\\ DRLN\\ PACN\\ TAN\end{tabular} &
  \begin{tabular}[c]{@{}c@{}}$\times 3$\\ $\times 3$\\ $\times 3$\\ $\times 3$\\ $\times 3$\\ $\times 3$\\ $\times 3$\\ $\times 3$\\ $\times 3$\\ $\times 3$\\ $\times 3$\\ $\times 3$\end{tabular} &
      \begin{tabular}[c]{@{}c@{}}-\\ 57K\\ 12K\\ 812K\\ 40.73M\\ 22.12M\\ 15.44M\\ 15.71M\\ 41.80M\\ 34.43M\\ 15.32M\\ 16.12M\end{tabular} &
%    \begin{tabular}[c]{@{}c@{}}-\\ -\\ -\\ -\\ 43.68M\\ 22.31M\\ 15.63M\\ 15.90M\\ 44.75M\\ 34.61M\end{tabular} &
%      \begin{tabular}[c]{@{}c@{}}$\times 3$\\ $\times 3$\\ $\times 3$\\ $\times 3$\\ $\times 3$\\ $\times 3$\end{tabular} &
  \multicolumn{1}{c|}{\begin{tabular}[c]{@{}c@{}}30.39 \\33.18\\33.67\\33.82\\ 34.65\\ 34.71\\ 34.74\\ 34.75\\ \textbf{34.85}\\ 34.78\\ 34.80\\ 34.79\end{tabular}} &
  \begin{tabular}[c]{@{}c@{}}0.8682\\0.9140\\0.9210\\0.9227\\ 0.9280\\ 0.9296\\ 0.9299\\ 0.9300\\ \textbf{0.9306}\\ 0.9303\\ 0.9296\\ 0.9301\end{tabular} &
  \multicolumn{1}{c|}{\begin{tabular}[c]{@{}c@{}}27.55\\29.37\\29.78\\29.87\\ 30.52\\ 30.57\\ 30.65\\ 30.59\\ \underline{30.70}\\ \textbf{30.73}\\ 30.63\\ 30.66\end{tabular}} &
  \begin{tabular}[c]{@{}c@{}}0.7742\\0.8240\\0.8320\\0.8320\\ 0.8462\\ 0.8468\\ 0.8482\\ 0.8476\\ 0.8485\\ \textbf{0.8488}\\ 0.8480\\ 0.8483\end{tabular} &
  \multicolumn{1}{c|}{\begin{tabular}[c]{@{}c@{}}27.21\\28.53\\28.83\\28.82\\ 29.25\\ 29.26\\ 29.32\\ 29.33\\ 29.34\\ \textbf{29.36}\\ 29.30\\ \underline{29.35}\end{tabular}} &
  \begin{tabular}[c]{@{}c@{}}0.7385\\0.7910\\0.7990\\0.7980\\ 0.8093\\ 0.8093\\ 0.8111\\ 0.8112\\ \underline{0.8117}\\ \underline{0.8117}\\ 0.8108\\ \underline{0.8117}\end{tabular} &
  \multicolumn{1}{c|}{\begin{tabular}[c]{@{}c@{}}24.46\\26.43\\27.14\\27.07\\ 28.80\\ 28.80\\ 29.09\\ 28.93\\ \underline{29.25}\\ 29.21\\ 29.01\\ 29.15\end{tabular}} &
  \begin{tabular}[c]{@{}c@{}}0.7349\\0.8080\\0.8290\\0.8280\\ 0.8653\\ 0.8653\\ 0.8702\\ 0.8671\\ \underline{0.8726}\\ 0.8722\\ 0.8691\\ 0.8717\end{tabular} &
  \multicolumn{1}{c|}{\begin{tabular}[c]{@{}c@{}}26.95\\31.10\\32.01\\32.21\\ 34.17\\ 34.13\\ 34.44\\ 34.30\\ 34.57\\ \textbf{34.71}\\ 34.45\\ 34.59\end{tabular}} &
  \begin{tabular}[c]{@{}c@{}}0.8556\\0.9210\\0.9340\\0.9350\\ 0.9476\\ 0.9484\\ 0.9499\\ 0.9494\\ 0.9508\\ \underline{0.9509}\\ 0.9497\\ 0.9502\end{tabular} \\ \hline
\begin{tabular}[c]{@{}c@{}}Backbone\\ LCoAN\end{tabular} &
  \begin{tabular}[c]{@{}c@{}}$\times 3$\\ $\times 3$\end{tabular} &
      \begin{tabular}[c]{@{}c@{}}14.03M\\ 15.67M\end{tabular} &
%    \begin{tabular}[c]{@{}c@{}}14.77M\\ 16.41M\end{tabular} &
%      \begin{tabular}[c]{@{}c@{}}$\times 3$\\ $\times 3$\end{tabular} &
  \multicolumn{1}{c|}{\begin{tabular}[c]{@{}c@{}} 34.67 \\ \textbf{34.85}\end{tabular}} &
  \begin{tabular}[c]{@{}c@{}} 0.9292 \\ \underline{0.9304}\end{tabular} &
  \multicolumn{1}{c|}{\begin{tabular}[c]{@{}c@{}} 30.53 \\ 30.69\end{tabular}} &
  \begin{tabular}[c]{@{}c@{}} 0.8464 \\ \underline{0.8487}\end{tabular} &
  \multicolumn{1}{c|}{\begin{tabular}[c]{@{}c@{}} 29.26 \\ \underline{29.35}\end{tabular}} &
  \begin{tabular}[c]{@{}c@{}}\ 0.8100 \\ \textbf{0.8122}\end{tabular} &
  \multicolumn{1}{c|}{\begin{tabular}[c]{@{}c@{}} 29.26 \\ \textbf{29.28}\end{tabular}} &
  \begin{tabular}[c]{@{}c@{}} 0.8657 \\ \textbf{0.8737}\end{tabular} &
  \multicolumn{1}{c|}{\begin{tabular}[c]{@{}c@{}} 34.17 \\ \underline{34.68}\end{tabular}} &
  \begin{tabular}[c]{@{}c@{}} 0.9486 \\ \textbf{0.9512}\end{tabular} \\ \hline
\begin{tabular}[c]{@{}c@{}}Bicubic\\ FSRCNN\\ VDSR\\ LapSRN\\ EDSR\\ RDN\\ RCAN\\ SAN\\ NLSN\\ DRLN \\ PACN\\ TAN\end{tabular} &
  \begin{tabular}[c]{@{}c@{}}$\times 4$\\ $\times 4$\\ $\times 4$\\ $\times 4$\\ $\times 4$\\ $\times 4$\\ $\times 4$\\ $\times 4$\\ $\times 4$\\ $\times 4$\\ $\times 4$\\ $\times 4$\end{tabular} &
      \begin{tabular}[c]{@{}c@{}}-\\ 57K\\ 12K\\ 812K\\ 40.73M\\ 22.12M\\ 15.44M\\ 15.71M\\ 41.80M\\ 34.43M\\ 15.32M\\ 16.12M\end{tabular} &
%    \begin{tabular}[c]{@{}c@{}}-\\ -\\ -\\ -\\ 43.09M\\ 22.27M\\ 15.59M\\ 15.86M\\ 44.16M\\ 34.58M\end{tabular} &
%      \begin{tabular}[c]{@{}c@{}}$\times 4$\\ $\times 4$\\ $\times 4$\\ $\times 4$\\ $\times 4$\\ $\times 4$\end{tabular} &
  \multicolumn{1}{c|}{\begin{tabular}[c]{@{}c@{}}28.42\\30.72\\31.35\\31.54\\ 32.46\\ 32.47\\ 32.63\\ \underline{32.64}\\ 32.59\\ 32.63\\ 32.56\\ 32.63\end{tabular}} &
  \begin{tabular}[c]{@{}c@{}}0.8104\\0.8660\\0.8830\\0.8850\\ 0.8968\\ 0.8990\\ \underline{0.9002}\\ \textbf{0.9003}\\ 0.9000\\ \underline{0.9002}\\ 0.8989\\ 0.9001\end{tabular} &
  \multicolumn{1}{c|}{\begin{tabular}[c]{@{}c@{}}26.00\\27.61\\28.02\\28.19\\ 28.80\\ 28.81\\ 28.87\\ \underline{28.92}\\ 28.87\\ \textbf{28.94}\\ 28.88\\ 28.90\end{tabular}} &
  \begin{tabular}[c]{@{}c@{}}0.7027\\0.7550\\0.7680\\0.7720\\ 0.7876\\ 0.7871\\ 0.7889\\ 0.7888\\ 0.7891\\ \textbf{0.7900}\\ 0.7886\\ 0.7892\end{tabular} &
  \multicolumn{1}{c|}{\begin{tabular}[c]{@{}c@{}}25.96\\26.98\\27.29\\27.32\\ 27.71\\ 27.72\\ 27.77\\ 27.78\\ 27.78\\ \textbf{27.83}\\ 27.76\\ 27.80\end{tabular}} &
  \begin{tabular}[c]{@{}c@{}}0.6675\\0.7150\\0.0726\\0.7270\\ 0.7420\\ 0.7419\\ 0.7436\\ 0.7436\\ 0.7444\\ 0.7444\\ 0.7432\\ \underline{0.7445}\end{tabular} &
  \multicolumn{1}{c|}{\begin{tabular}[c]{@{}c@{}}23.14\\24.62\\25.18\\25.21\\ 26.64\\ 26.61\\ 26.82\\ 26.79\\ 26.96\\ \underline{26.98}\\ 26.84\\ 26.84\end{tabular}} &
  \begin{tabular}[c]{@{}c@{}}0.6577\\0.7280\\0.7540\\0.7560\\ 0.8033\\ 0.8028\\ 0.8087\\ 0.8068\\ 0.8109\\ \underline{0.8119}\\ 0.8087\\ 0.8094\end{tabular} &
  \multicolumn{1}{c|}{\begin{tabular}[c]{@{}c@{}}24.89\\27.90\\28.83\\29.09\\ 31.02\\ 31.00\\ 31.22\\ 31.18\\ 31.27\\ \textbf{31.54}\\ 31.33\\ 31.46\end{tabular}} &
  \begin{tabular}[c]{@{}c@{}}0.7866\\0.8610\\0.8870\\0.8900\\ 0.9148\\ 0.9151\\ 0.9173\\ 0.9169\\ 0.9184\\ \underline{0.9196}\\ 0.9178\\ 0.9184\end{tabular} \\ \hline
\begin{tabular}[c]{@{}c@{}}Backbone\\ LCoAN\end{tabular} &
  \begin{tabular}[c]{@{}c@{}}$\times 4$\\ $\times 4$\end{tabular} &
      \begin{tabular}[c]{@{}c@{}}14.03M\\ 15.67M\end{tabular} &
%    \begin{tabular}[c]{@{}c@{}}14.62M\\ 16.26M\end{tabular} &
%      \begin{tabular}[c]{@{}c@{}}$\times 4$\\ $\times 4$\end{tabular} &
  \multicolumn{1}{c|}{\begin{tabular}[c]{@{}c@{}} 32.46 \\ \textbf{32.65} \end{tabular}} &
  \begin{tabular}[c]{@{}c@{}} 0.8976 \\ 0.8999 \end{tabular} &
  \multicolumn{1}{c|}{\begin{tabular}[c]{@{}c@{}} 28.77 \\ 28.91\end{tabular}} &
  \begin{tabular}[c]{@{}c@{}} 0.7871 \\ \underline{0.7896}\end{tabular} &
  \multicolumn{1}{c|}{\begin{tabular}[c]{@{}c@{}} 27.71 \\ \underline{27.79}\end{tabular}} &
  \begin{tabular}[c]{@{}c@{}} 0.7423 \\ \textbf{0.7452}\end{tabular} &
  \multicolumn{1}{c|}{\begin{tabular}[c]{@{}c@{}} 26.59 \\ \textbf{27.02}\end{tabular}} &
  \begin{tabular}[c]{@{}c@{}} 0.8022  \\ \textbf{0.8132}\end{tabular} &
  \multicolumn{1}{c|}{\begin{tabular}[c]{@{}c@{}} 31.05 \\ \underline{31.48}\end{tabular}} &
  \begin{tabular}[c]{@{}c@{}} 0.9155 \\ \textbf{0.9200}\end{tabular} \\ \hline
\end{tabular}
}
\end{center}
\end{table*}

% todo 消融实验
\begin{figure}[!htpb]
	\hspace{-6mm}
	\centering
	\subfloat[]{
		\label{fig_plot_cluster}
		\includegraphics[width=0.5\linewidth]{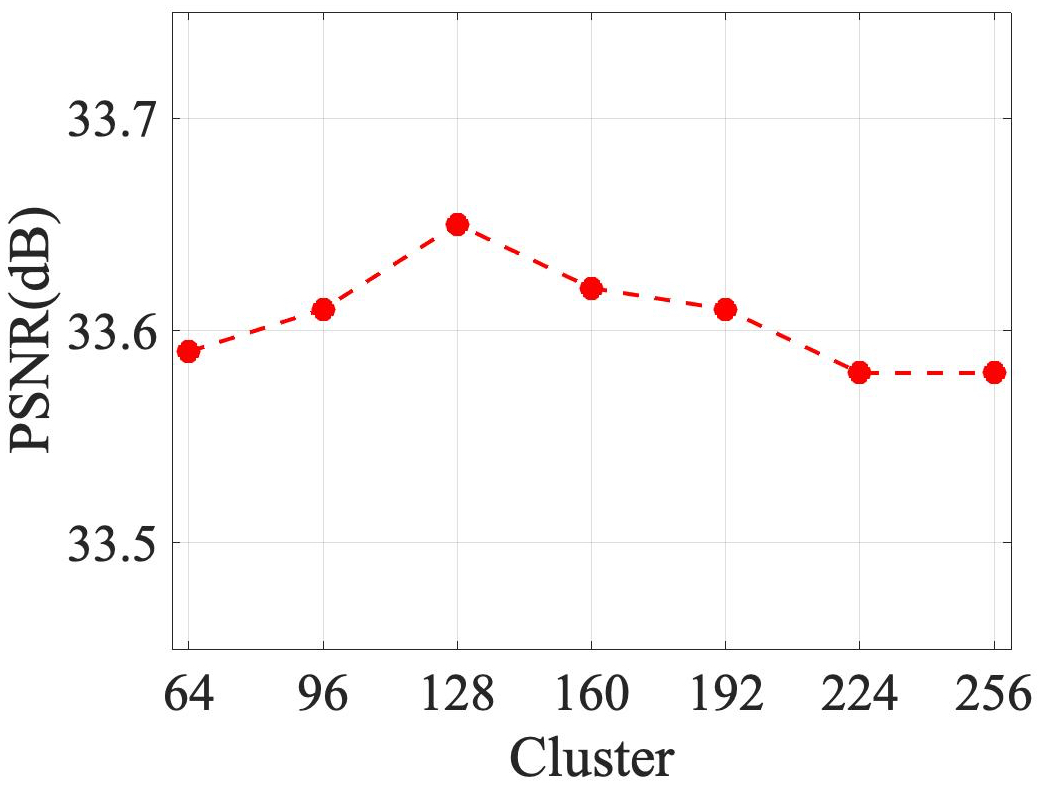}}
	\subfloat[]{
		\label{fig_plot_window_size}
		\includegraphics[width=0.5\linewidth]{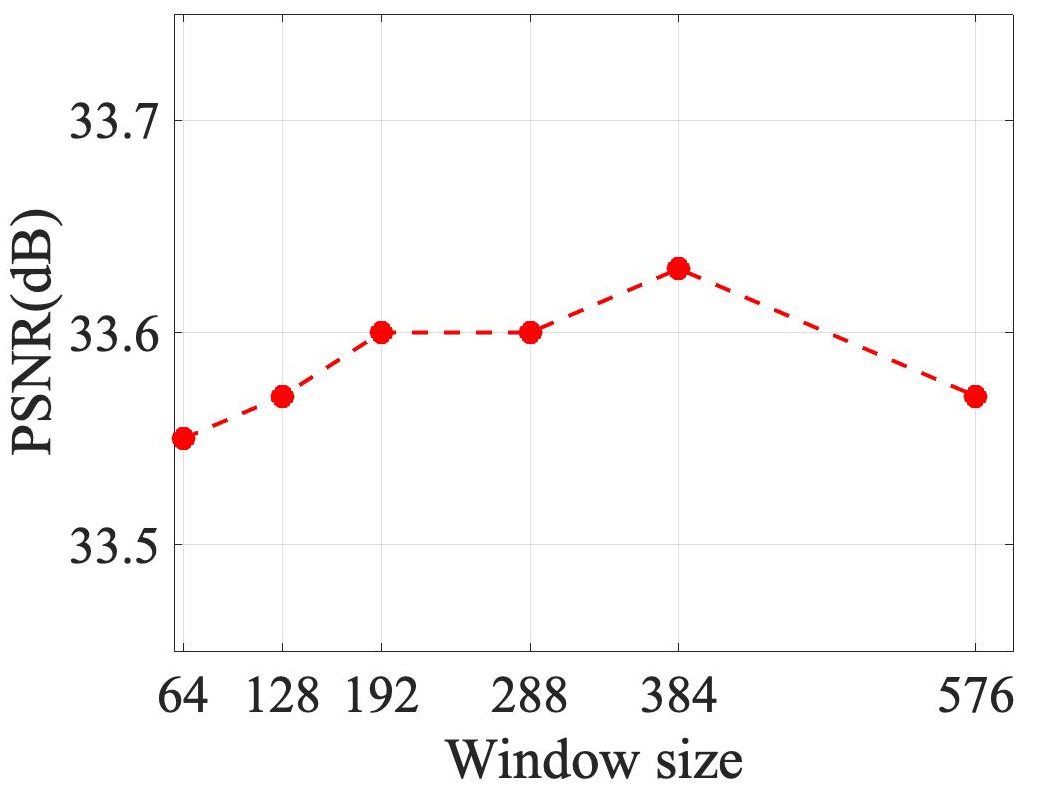}}
	\caption{Ablation experiments conducted on Set14 with scale factor 2 to explore the effects of cluster and window size. (a) The PSNR results from different cluster settings. (b) The PSNR results from different window size settings.}
\end{figure}

\subsubsection{Cluster and Window Size}
In LSP, we use \emph{k}-means algorithm to cluster features to explore the sparse prior of natural images. When the $k$ is too small, it may cause large differences within clusters and small differences between clusters, affecting the clustering performance. Similarly, when the $k$ value is too large, it may result in small differences within clusters and large differences between clusters, also affecting the clustering performance. The effects of different k values are shown in \cref{fig_plot_cluster}. Our LSP achieves the best SR performance when $k = 128$.

% m109图片
\begin{figure*}[!htbp]
	
	\begin{minipage}[]{0.48\linewidth}
		\centering
		    \begin{minipage}[]{0.46\linewidth}
        \centering
        \setlength{\abovecaptionskip}{1.25mm}
        \includegraphics[height=1.15\linewidth,width=\linewidth]{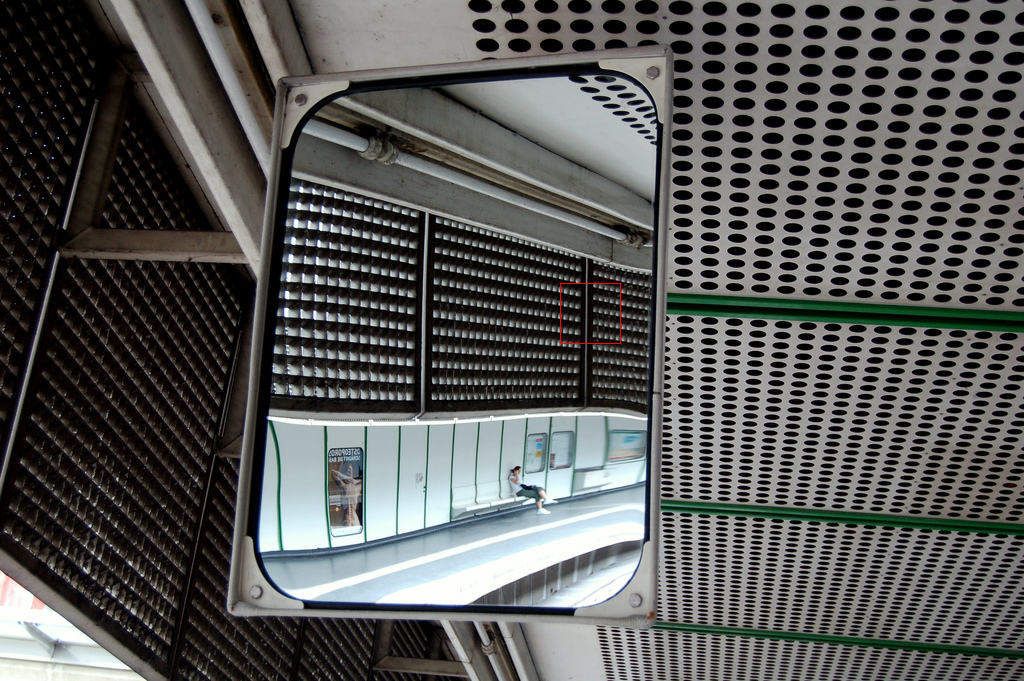}
        \caption*{img004}
    \end{minipage}
    \begin{minipage}[]{0.23\linewidth}
        \centering
        	
        	\begin{subfigure}[]{\linewidth}
				\centering
				\includegraphics[height=\linewidth,width=\linewidth]{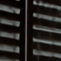}
        		\caption*{EDSR}
        		
        		\includegraphics[height=\linewidth,width=\linewidth]{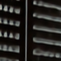}
        		\caption*{RCAN}
			\end{subfigure}
    \end{minipage}
    \begin{minipage}[]{0.23\linewidth}
        \centering
        	
        	\begin{subfigure}[]{\linewidth}
				\centering
				\includegraphics[height=\linewidth,width=\linewidth]{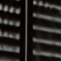}
        		\caption*{RDN}
        		
        		\includegraphics[height=\linewidth,width=\linewidth]{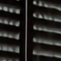}
        		\caption*{NLSN}
			\end{subfigure}
    \end{minipage}
    \begin{minipage}[]{0.23\linewidth}
        \centering
        	
        	\begin{subfigure}[]{\linewidth}
				\centering
				\includegraphics[height=\linewidth,width=\linewidth]{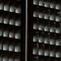}
        		\caption*{HR}
			\end{subfigure}
    \end{minipage}
    \begin{minipage}[]{0.23\linewidth}
        \centering
        	
        	\begin{subfigure}[]{\linewidth}
				\centering
				\includegraphics[height=\linewidth,width=\linewidth]{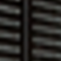}
        		\caption*{Bicubic}
			\end{subfigure}
    \end{minipage}
    \begin{minipage}[]{0.23\linewidth}
        \centering
        	
        	\begin{subfigure}[]{\linewidth}
				\centering
				\includegraphics[height=\linewidth,width=\linewidth]{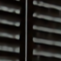}
        		\caption*{DRLN}
			\end{subfigure}
    \end{minipage}
    \begin{minipage}[]{0.23\linewidth}
        \centering
        	
        	\begin{subfigure}[]{\linewidth}
				\centering
				\includegraphics[height=\linewidth,width=\linewidth]{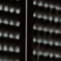}
        		\caption*{LCoAN(ours)}
			\end{subfigure}
    \end{minipage}
    \centering
	\end{minipage}
	\begin{minipage}[]{0.48\linewidth}
		\centering
		    \begin{minipage}[]{0.46\linewidth}
        \centering
        \setlength{\abovecaptionskip}{1.25mm}
        \includegraphics[height=1.15\linewidth,width=\linewidth]{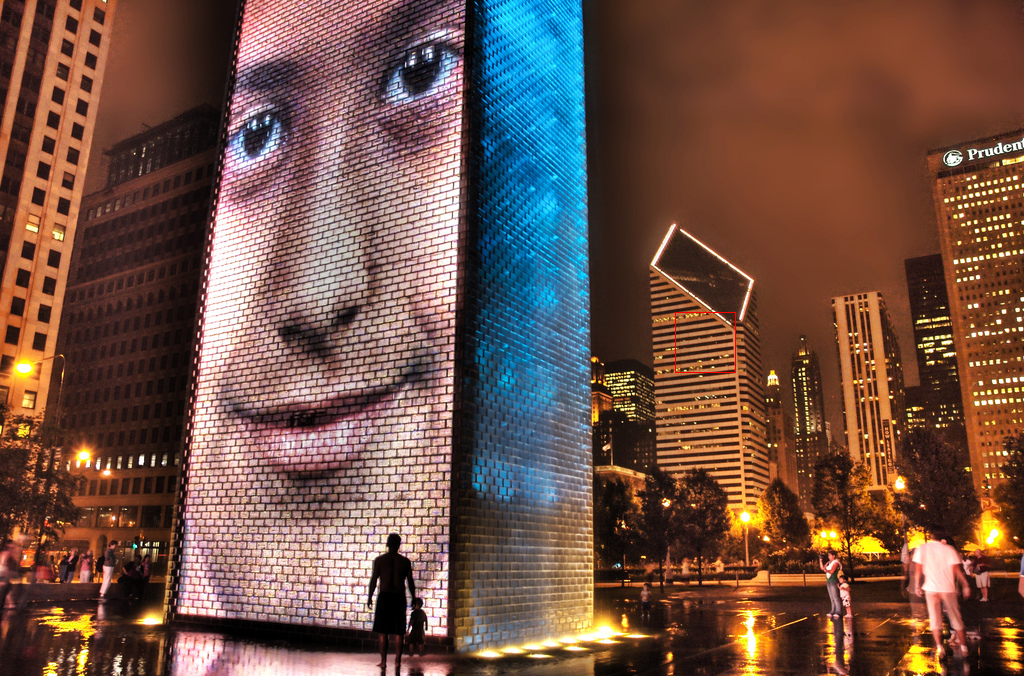}
        \caption*{img076}
    \end{minipage}
    \begin{minipage}[]{0.23\linewidth}
        \centering
        	
        	\begin{subfigure}[]{\linewidth}
				\centering
				\includegraphics[height=\linewidth,width=\linewidth]{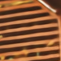}
        		\caption*{EDSR}
        		
%        		\vspace{1mm}
        		\includegraphics[height=\linewidth,width=\linewidth]{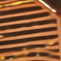}
        		\caption*{RCAN}
			\end{subfigure}
    \end{minipage}
    \begin{minipage}[]{0.23\linewidth}
        \centering
        	
        	\begin{subfigure}[]{\linewidth}
				\centering
				\includegraphics[height=\linewidth,width=\linewidth]{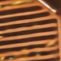}
        		\caption*{RDN}
        		
%        		\vspace{1mm}
        		\includegraphics[height=\linewidth,width=\linewidth]{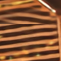}
        		\caption*{NLSN}
			\end{subfigure}
    \end{minipage}
    \centering
    
    \begin{minipage}[]{0.23\linewidth}
        \centering
        	
        	\begin{subfigure}[]{\linewidth}
				\centering
				\includegraphics[height=\linewidth,width=\linewidth]{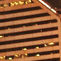}
        		\caption*{HR}
			\end{subfigure}
    \end{minipage}
    \begin{minipage}[]{0.23\linewidth}
        \centering
        	
        	\begin{subfigure}[]{\linewidth}
				\centering
				\includegraphics[height=\linewidth,width=\linewidth]{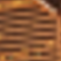}
        		\caption*{Bicubic}
			\end{subfigure}
    \end{minipage}
    \begin{minipage}[]{0.23\linewidth}
        \centering
        	
        	\begin{subfigure}[]{\linewidth}
				\centering
				\includegraphics[height=\linewidth,width=\linewidth]{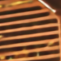}
        		\caption*{DRLN}
			\end{subfigure}
    \end{minipage}
    \begin{minipage}[]{0.23\linewidth}
        \centering
        	
        	\begin{subfigure}[]{\linewidth}
				\centering
				\includegraphics[height=\linewidth,width=\linewidth]{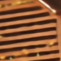}
        		\caption*{LCoAN(ours)}
			\end{subfigure}
    \end{minipage}
    \centering
	\end{minipage}
	\centering

\vspace{2mm}
   \caption{Visual comparisons on Urban100($\times 4$).}
   \label{fig_visual_comparison_u100}
\end{figure*}

% urban100图片
\begin{figure*}[!htbp]
	
	\begin{minipage}[]{0.48\linewidth}
		\centering
		    \begin{minipage}[]{0.46\linewidth}
        \centering
        \setlength{\abovecaptionskip}{1.25mm}
        \includegraphics[height=1.15\linewidth,width=\linewidth]{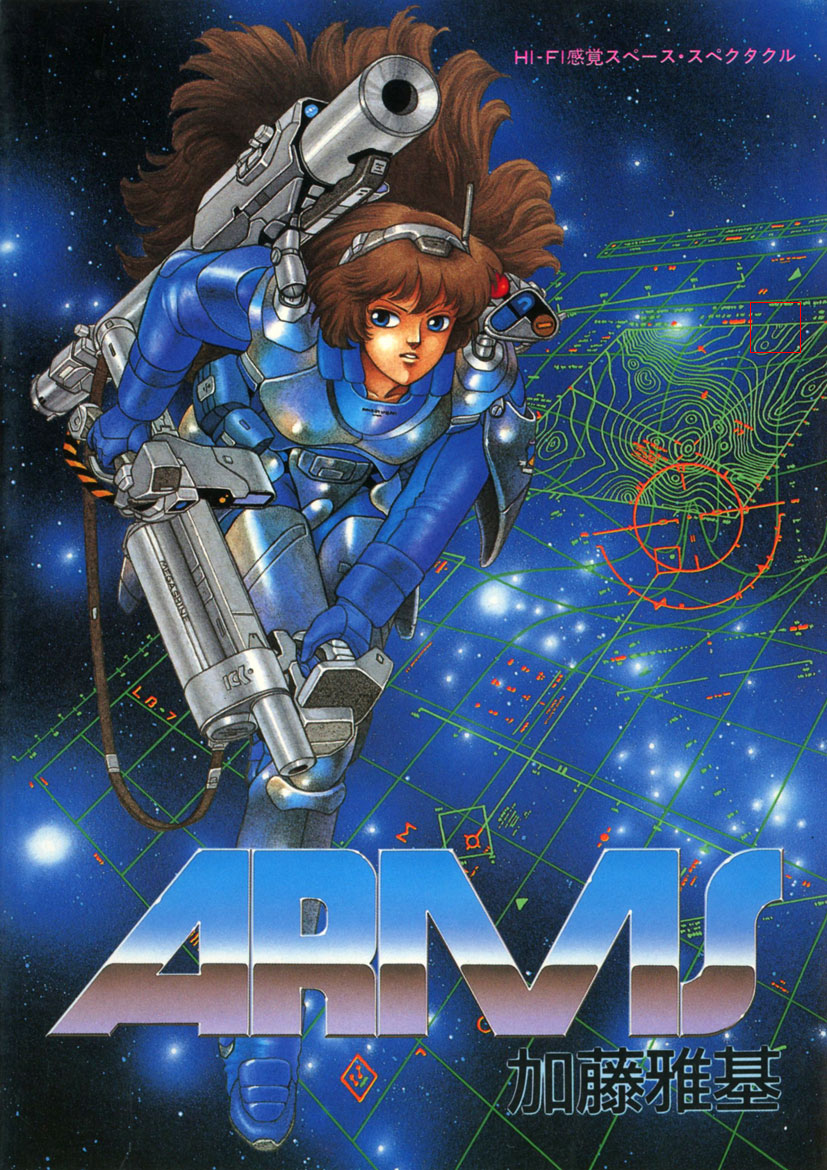}
        \caption*{ARMS}
    \end{minipage}
    \begin{minipage}[]{0.23\linewidth}
        \centering
        	
        	\begin{subfigure}[]{\linewidth}
				\centering
				\includegraphics[height=\linewidth,width=\linewidth]{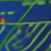}
        		\caption*{EDSR}
        		
        		\includegraphics[height=\linewidth,width=\linewidth]{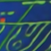}
        		\caption*{RCAN}
			\end{subfigure}
    \end{minipage}
    \begin{minipage}[]{0.23\linewidth}
        \centering
        	
        	\begin{subfigure}[]{\linewidth}
				\centering
				\includegraphics[height=\linewidth,width=\linewidth]{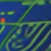}
        		\caption*{RDN}
        		
        		\includegraphics[height=\linewidth,width=\linewidth]{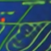}
        		\caption*{NLSN}
			\end{subfigure}
    \end{minipage}
    \begin{minipage}[]{0.23\linewidth}
        \centering
        	
        	\begin{subfigure}[]{\linewidth}
				\centering
				\includegraphics[height=\linewidth,width=\linewidth]{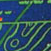}
        		\caption*{HR}
			\end{subfigure}
    \end{minipage}
    \begin{minipage}[]{0.23\linewidth}
        \centering
        	
        	\begin{subfigure}[]{\linewidth}
				\centering
				\includegraphics[height=\linewidth,width=\linewidth]{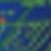}
        		\caption*{Bicubic}
			\end{subfigure}
    \end{minipage}
    \begin{minipage}[]{0.23\linewidth}
        \centering
        	
        	\begin{subfigure}[]{\linewidth}
				\centering
				\includegraphics[height=\linewidth,width=\linewidth]{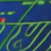}
        		\caption*{DRLN}
			\end{subfigure}
    \end{minipage}
    \begin{minipage}[]{0.23\linewidth}
        \centering
        	
        	\begin{subfigure}[]{\linewidth}
				\centering
				\includegraphics[height=\linewidth,width=\linewidth]{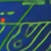}
        		\caption*{LCoAN(ours)}
			\end{subfigure}
    \end{minipage}
    \centering
	\end{minipage}
	\begin{minipage}[]{0.48\linewidth}
		\centering
		    \begin{minipage}[]{0.46\linewidth}
        \centering
        \setlength{\abovecaptionskip}{1.25mm}
        \includegraphics[height=1.15\linewidth,width=\linewidth]{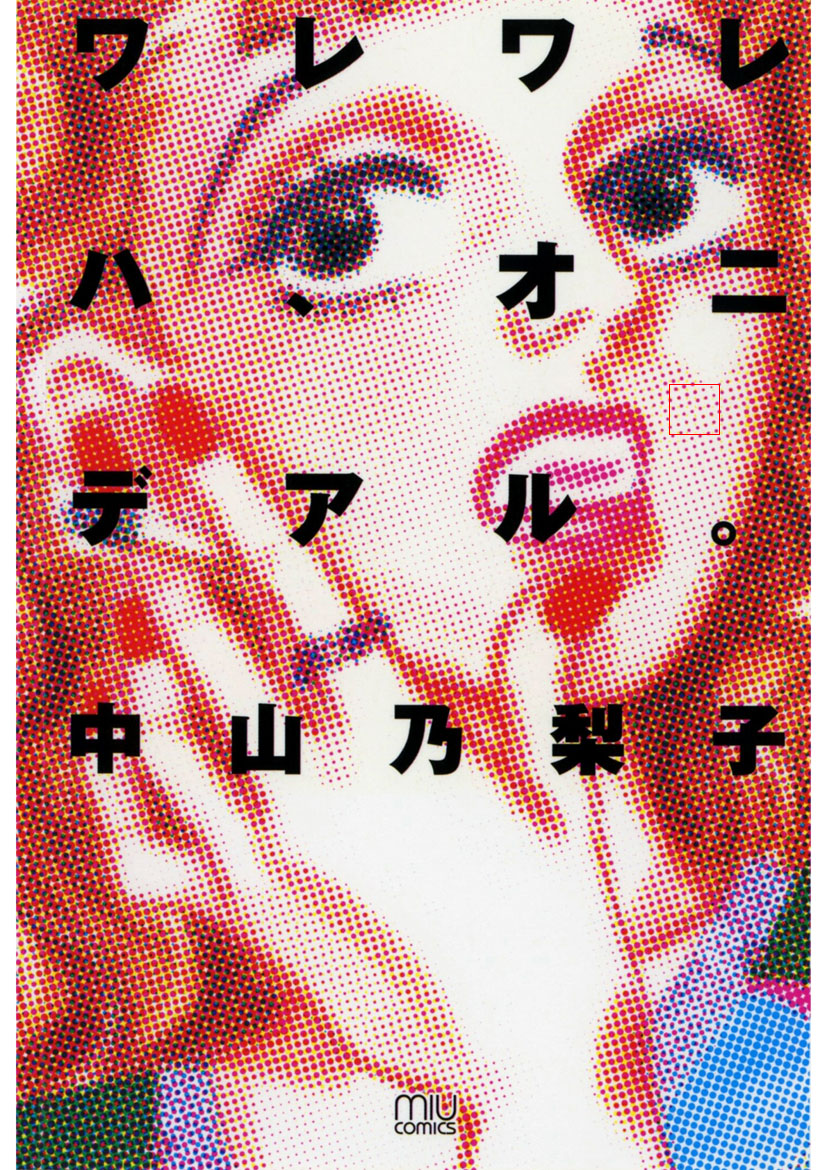}
        \caption*{WarewareHaOniDearu}
    \end{minipage}
    \begin{minipage}[]{0.23\linewidth}
        \centering
        	
        	\begin{subfigure}[]{\linewidth}
				\centering
				\includegraphics[height=\linewidth,width=\linewidth]{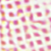}
        		\caption*{EDSR}
        		
%        		\vspace{1mm}
        		\includegraphics[height=\linewidth,width=\linewidth]{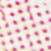}
        		\caption*{RCAN}
			\end{subfigure}
    \end{minipage}
    \begin{minipage}[]{0.23\linewidth}
        \centering
        	
        	\begin{subfigure}[]{\linewidth}
				\centering
				\includegraphics[height=\linewidth,width=\linewidth]{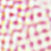}
        		\caption*{RDN}
        		
%        		\vspace{1mm}
        		\includegraphics[height=\linewidth,width=\linewidth]{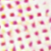}
        		\caption*{NLSN}
			\end{subfigure}
    \end{minipage}
    \centering
    
    \begin{minipage}[]{0.23\linewidth}
        \centering
        	
        	\begin{subfigure}[]{\linewidth}
				\centering
				\includegraphics[height=\linewidth,width=\linewidth]{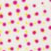}
        		\caption*{HR}
			\end{subfigure}
    \end{minipage}
    \begin{minipage}[]{0.23\linewidth}
        \centering
        	
        	\begin{subfigure}[]{\linewidth}
				\centering
				\includegraphics[height=\linewidth,width=\linewidth]{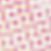}
        		\caption*{Bicubic}
			\end{subfigure}
    \end{minipage}
    \begin{minipage}[]{0.23\linewidth}
        \centering
        	
        	\begin{subfigure}[]{\linewidth}
				\centering
				\includegraphics[height=\linewidth,width=\linewidth]{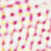}
        		\caption*{DRLN}
			\end{subfigure}
    \end{minipage}
    \begin{minipage}[]{0.23\linewidth}
        \centering
        	
        	\begin{subfigure}[]{\linewidth}
				\centering
				\includegraphics[height=\linewidth,width=\linewidth]{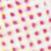}
        		\caption*{LCoAN(ours)}
			\end{subfigure}
    \end{minipage}
    \centering
	\end{minipage}
	\centering

\vspace{2mm}
   \caption{Visual comparisons on Manga($\times 4$).}
   \label{fig_visual_comparison_m109}
\end{figure*}

As discussed in \cref{sec_detail_lca}, the window size determines the number of non-local features that can be explored by the query feature. The impact of different window sizes is shown in \cref{fig_plot_window_size}, from which we can see that our LCoA achieves peak SR performance when the window size is set to 384. As the window size increases further, the performance of SR will start to decline. This is because a larger window size spans multiple clustered features, which reduces the performance gain from sparsity. Conversely, a window size that is too small may lead to insufficient generalization ability of the weight sharing strategy, resulting in a decrease in the effectiveness of LCoA.  Therefore, when choosing the window size, it is necessary to balance the trade-off between sparsity and generalization ability in order to achieve better SR performance.
% As the window size increases further, the performance of SR will begin to decline. This is because a window size that is too small can lead to insufficient generalization ability of the weight sharing strategy, thus reducing the effectiveness of LRA. Conversely, a larger window size that spans multiple clustering features can also reduce the performance improvement brought by sparsity.

\subsection{Comparisons with State-of-the-art}
To demonstrate the effectiveness of our Learnable Collaborative Attention (LCoA), we compare LCoAN with 11 state-of-the-art convolutional-based models including FSRCNN\cite{dong2016accelerating}, VDSR\cite{kim2016accurate}, LapSRN\cite{lai2017deep}, EDSR \cite{lim2017enhanced}, RDN \cite{zhang2018residual}, RCAN \cite{zhang2018image}, SAN \cite{dai2019second}, NLSN \cite{mei2021image}, DRLN \cite{saeed2022densely}, PACN \cite{wang2023pixel}, and TAN\cite{yang2023single}.

The quantitative results are shown in \cref{tab:x2_x3_x4_psnr_ssim}. We can see that compared to other state-of-the-art deep image SR models, our LCoAN demonstrates competitive performance on all benchmarks and scale factors. Compared to the backbone network, adding LCoA has shown great advantages in performance improvement, and even exceeded the highly competitive NLSN in performance. For scale factor 2, the proposed LCoAN brings about 0.1dB improvement in Set5 and B100, 0.2dB improvement in Set14, and over 0.4dB improvement in Urban100 and Manga109. These performance gains indicate that LCoA has successfully explored the self-similarity prior of natural images for more accurate super-resolution. The qualitative evaluations on Urban100 and Manga109 are shown in \cref{fig_visual_comparison_u100} and \cref{fig_visual_comparison_m109}, respectively. From the visual comparison results, it can be seen that our LCoAN generates visually appealing results with accurate image textures. These results indicate that our LCoAN achieves competitive performance in both quantitative metrics and perceptual quality compared to other deep SR models.

\section{Conclusion}
%The present work conducts an in-depth analysis of the Non-Local Attention (NLA) technique and reveals its critical limitations for SISR tasks. We propose a novel approach, Decoupled Non-Local Attention (DNLA), to overcome these limitations and facilitate the effective and efficient aggregation of highly related features in deep networks. By leveraging FlyHash to partition image features, DNLA can globally identify the most valuable locations to attend without allocating attention to unrelated regions, thereby improving computational efficiency. Our proposed DNLA is integrated into a residual backbone architecture and sets new state-of-the-art performance benchmarks across multiple evaluation metrics. Extensive evaluations demonstrate that our DNLA outperforms NLA and provides substantial benefits for achieving accurate image super-resolution.
In this paper, we propose an efficient Learnable Collaborative Attention (LCoA) to improve the computational efficiency of non-local modeling in SR tasks. The LCoA comprises two parts, namely the Learnable Sparse Pattern (LSP) that enforces learnable sparsity constraints, and the Collaborative Attention (CoA) that applies weight sharing to the non-local modeling process. By introducing learnable sparsity and weight sharing biases into non-local operation, our LCoA exhibits a significant computational efficiency advantage and achieves competitive SR performance. Experimental results on several popular datasets confirm the values of LSP and CoA, demonstrate the superiority of our LCoA over representative efficient attention methods.

\bibliographystyle{IEEEtranS}
\bibliography{egbib}

\end{document}